\documentclass{article} 
\usepackage{iclr2027_conference,times}
\usepackage{natbib}
\usepackage{graphicx}
\usepackage{booktabs}
\usepackage{multirow}
\usepackage{makecell}
\usepackage[table]{xcolor}

\usepackage{amsmath,amsfonts,bm}

\def\eqref#1{equation~\ref{#1}}

\def\1{\bm{1}}

\DeclareMathAlphabet{\mathsfit}{\encodingdefault}{\sfdefault}{m}{sl}
\SetMathAlphabet{\mathsfit}{bold}{\encodingdefault}{\sfdefault}{bx}{n}

\usepackage{fontawesome5}
\usepackage{hyperref}
\usepackage{url}
\usepackage{caption}
\usepackage{wrapfig}
\usepackage{xcolor}
\usepackage{pifont}
\newcommand{\cmark}{\textcolor{green!60!black}{\ding{51}}}
\newcommand{\xmark}{\textcolor{red!70!black}{\ding{55}}}

\title{Scaling Versatile 3D Assets Editing with a Million-Scale Dataset}
\author{Badi Li\textsuperscript{\normalfont1,2,4}, Tianxin Huang\textsuperscript{\normalfont1}, Yu Zhou\textsuperscript{\normalfont3}, Wei-Shi Zheng\textsuperscript{\normalfont2,4}, Yi Ma\textsuperscript{\normalfont1,2}, Shenghua Gao\textsuperscript{\normalfont1,2}\thanks{Correspondence to: gaosh@hku.hk}  \\ \\
\textsuperscript{1} The University of Hong Kong
~\textsuperscript{2} Shenzhen Loop Area Institute \\
\textsuperscript{3} Shanghai Innovation Institute
~\textsuperscript{4} Sun Yat-Sen University 
}

\usepackage{xcolor}

\definecolor{taskcolor1}{RGB}{220, 50, 50}    
\definecolor{taskcolor2}{RGB}{30, 100, 200}   
\definecolor{taskcolor3}{RGB}{40, 150, 70}    
\definecolor{taskcolor4}{RGB}{230, 150, 20}   
\definecolor{taskcolor5}{RGB}{200, 60, 120}   
\definecolor{black}{RGB}{0, 0, 0}

\iclrfinalcopy
\begin{document}

\maketitle

\faGlobe~\href{https://libd1.github.io/Alchemy3D-Project/}{\texttt{Project Page}} 
\hfill
\faGithub~\href{https://github.com/libd1/Alchemy3D}{\texttt{Code}}
\hfill
\faGithub~\href{https://github.com/libd1/edit3dstudio}{\texttt{Evaluation}}
\hfill
\faHuggingFace~\href{https://huggingface.co/libadi/Alchemy3D}{\texttt{Model}}
\hfill
\faHuggingFace~\href{https://huggingface.co/datasets/libadi/Alchemy3D-1M}{\texttt{Dataset}}
\hfill
\faHuggingFace~\href{https://huggingface.co/datasets/libadi/GEdit3D-Bench}{\texttt{Benchmark}}
\begin{figure}[h]
\begin{center}
\includegraphics[width=0.9\linewidth]{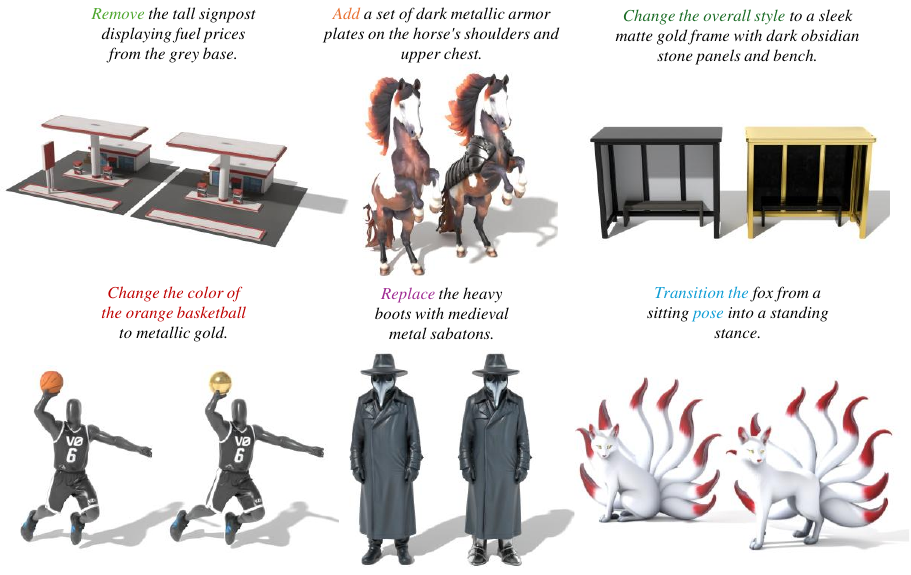}
\end{center}
\caption{
Given a 3D asset and an editing instruction, our method edits the asset accordingly while preserving unrelated regions.
}
\label{fig:teaser}
\end{figure}
\begin{abstract}
    Although recent 3D generative models produce increasingly realistic assets, controllable 3D asset editing remains challenging. Existing methods are limited by scarce training data, insufficient source-aware modeling, and a lack of practical evaluation protocols.
To address these limitations, we present Alchemy3D, a unified framework for training and evaluating versatile 3D asset editors that covers data construction, model architecture, and benchmark evaluation. Specifically, we curate Alchemy3D-1M, a large-scale 3D editing dataset containing 1.25M assets and 1.38M editing pairs across seven editing types. On this data, we train a family of generative flow models for general-purpose 3D asset editing. The model family supports image- and text-conditioned editing, few-step inference, and transfer to multi-view 3D part segmentation. We further introduce GEdit3D-Bench, a large-scale, open-world benchmark with a multi-dimensional evaluation protocol. Across existing and newly introduced benchmarks, our method outperforms prior methods on most metrics of editing fidelity, source preservation, and visual quality.

\end{abstract}

\section{Introduction}


With the development of diffusion and flow-matching models~\citep{ncsn,score-sde,ddpm,flowmatching}, recent 3D generative models~\citep{trellis,trellis2,hunyuan3d2.1} can produce high-quality 3D assets from text or image prompts. However, general-purpose 3D editing remains comparatively underdeveloped because it requires fine-grained spatial awareness. An editing model must not only execute a localized or global modification according to the instruction, but also keep irrelevant regions unchanged.

Compared to generation methods~\citep{trellis,trellis2,hunyuan3d2.1}, existing 3D editing models are typically constrained by limitations in training data, model design, and evaluation protocols. First, achieving region-specific revisions requires paired data before and after editing, which is difficult to acquire. Existing 3D editing datasets are limited in scale, diversity, quality, and editing types~\citep{steer3d, 3deditformer,partflow}.
Second, recent 3D editors~\citep{steer3d,partflow} inject the source asset through a ControlNet-style~\citep{controlnet} branch, which limits the interaction between source and target and may provide insufficient fine-grained guidance from the source.
Finally, existing editing benchmarks are often relatively small, drawn from the same distribution as the training data, or evaluated by similarity to a single automatically generated target~\citep{voxhammer,anchorflow,nano3d,3deditformer,partflow}, which restricts evaluation to in-distribution settings and fails to capture the one-to-many nature of generative editing.

To address these challenges, we present \textbf{Alchemy3D}, an integrated 3D editing framework comprising scalable data construction, model architecture design, and benchmark evaluation. For data construction, we select seven representative editing types (addition, removal, replacement, animation, local appearance, global appearance, and segmentation) and design type-specific pipelines to create editing pairs at scale. Candidate pairs are filtered with vision-language models~\citep{qwen3.6-27b,Qwen3-VL}, yielding 1.38 million editing pairs.

For the model architecture, we concatenate the tokens of the source asset and the edited target asset so that they interact through self-attention, while text or image instructions are injected through cross-attention. To improve inference efficiency, we further adopt a few-step distillation procedure based on MeanFlow~\citep{meanflow} and DMD~\citep{dmd,dmd2}. With this architecture and \textbf{Alchemy3D-1M}, we train a family of \textbf{Alchemy3D} model variants for different editing scenarios.

For evaluation, we construct \textbf{GEdit3D-Bench}, a benchmark built from assets that are independent of our training data. It combines newly synthesized 3D content with recently released real-world assets from Sketchfab\footnote{\url{https://sketchfab.com/}} to mitigate in-distribution evaluation. Editing instructions are written by a vision-language model, target images are produced by an image editing model~\citep{hunyuanimage3}, and all samples are curated through automated and human filtering.
Rather than measuring similarity to a single generated target, GEdit3D-Bench jointly assesses view quality, reference alignment, and Multimodal Large Language Model (MLLM)-based scores, providing a more comprehensive evaluation of 3D editors.

Our contributions are summarized as follows:
\begin{itemize}
    \item To expand the scope of versatile 3D editing, we propose a scalable data construction pipeline spanning 7 diverse editing tasks, yielding \textbf{Alchemy3D-1M}, a dataset of 1.38 million high-quality editing pairs covering diverse assets.
    \item Instead of introducing source asset features through a separate branch, we fuse tokens from the source and edited target models, enabling sufficient interaction between them. 
    We further propose a distillation pipeline that enables efficient, few-step inference for the 3D editing model. Building on this architecture and \textbf{Alchemy3D-1M}, we train a series of \textbf{Alchemy3D} model variants tailored to different editing use cases.
    \item To enable a more comprehensive and fair evaluation of 3D editors, we introduce \textbf{GEdit3D-Bench}, a large-scale, multi-dimensional benchmark independent of our training data.

    \item Extensive comparisons on multiple benchmarks confirm that our model achieves significant improvements over existing 3D editing methods, with the distilled model runs about $4\times$ faster than the base model while retaining comparable quality.
\end{itemize}

\section{Related Work}
\subsection{Native 3D Generation}
3D generative models synthesize representations such as point clouds~\citep{pointdiffusion,pointe}, neural fields~\citep{diffrf}, triplanes~\citep{rodin}, and 3D Gaussian splats~\citep{gvgen}. TRELLIS~\citep{trellis} introduced a structured latent space that supports meshes, NeRFs~\citep{nerf}, and 3D Gaussian splats~\citep{3dgs} through a unified VAE~\citep{vae}. TRELLIS.2~\citep{trellis2} subsequently introduced the compact O-Voxel representation and decomposed asset generation into sparse-structure, shape, and PBR-material stages. Alchemy3D adopts this representation and transforms the generation pipeline into a source-conditioned editor.

\subsection{3D Asset Editing}
Early 3D editing methods rely on optimization, either reconstructing assets from images modified by a 2D editor~\citep{instructnerf2nerf} or applying score-distillation sampling~\citep{dreamfusion} to optimize a 3D representation~\citep{voxe,focaldreamer,gsedit}. Subsequent systems edit multi-view renderings or videos and reconstruct an asset from the modified observations; agentic variants primarily automate view selection~\citep{tailor3d,edit360,editcast3d,pro3deditor}. Although flexible, these pipelines are computationally expensive and prone to cross-view inconsistency.

Feed-forward approaches improve efficiency by adapting techniques from zero-shot image editing. VoxHammer~\citep{voxhammer} combines TRELLIS~\citep{trellis} with RF-Inversion~\citep{rfinversion} and attention manipulation~\citep{rfsolver}, while Nano3D~\citep{nano3d} integrates TRELLIS with FlowEdit~\citep{flowedit}. Their reliance on training-free image-editing mechanisms, however, limits robustness on complex 3D transformations. Steer3D~\citep{steer3d} and 3DEditFormer~\citep{3deditformer} instead train 3D editors on constructed paired datasets, demonstrating the value of task-specific supervision but remaining constrained by data scale and coverage. Concurrently, PartFlow~\citep{partflow} constructs editing pairs from part-segmentation datasets and is the closest prior setting to ours. Alchemy3D extends this direction with substantially larger and broader dataset, a unified model family, and open-world evaluation.

\section{Alchemy3D-1M Dataset}
\label{sec:dataset}
\begin{table*}[t]
    \centering
    \caption{Comparison of existing 3D asset editing datasets.
    Our Alchemy3D-1M incorporates more diverse editing types and is more than 10$\times$ larger in scale than existing related datasets.}
    \label{tab:dataset_comparison}
    \small
    \setlength{\tabcolsep}{1.2pt}
    \renewcommand{\arraystretch}{0.95}
    \begin{tabular}{l@{\hspace{4pt}}cc@{\hspace{4pt}}cccccc}
        \toprule
        \textbf{Dataset}
        & \textbf{Scale}
        & \textbf{PBR}
        & \multicolumn{6}{c}{\textbf{Editing Types}} \\
        \cmidrule(lr){4-9}
        &
        &
        &
        \textbf{Add}
        & \textbf{Remove}
        & \textbf{Replace}
        & \textbf{Appearance}
        & \textbf{Animation}
        & \textbf{Segmentation} \\
        \midrule
        Steer3D
        & 100K
        & \xmark
        & \cmark & \cmark & \xmark & \cmark & \xmark & \xmark \\
        Nano3D-100K (ICLR'26)
        & 100K
        & \xmark
        & \cmark & \cmark & \cmark & \xmark & \xmark & \xmark \\
        3DEditVerse (ICML'26)
        & 116K
        & \xmark
        & \cmark & \cmark & \cmark & \xmark & \xmark & \xmark \\
        PxForm (SIGGRAPH Asia'26)
        & 102K
        & \xmark
        & \cmark & \cmark & \cmark & \cmark & \xmark & \xmark \\
        \textbf{Alchemy3D-1M}
        & \textbf{1.38M}
        & \cmark
        & \cmark & \cmark & \cmark & \cmark & \cmark & \cmark \\
        \bottomrule
    \end{tabular}
\end{table*}

\begin{figure}[h]
\begin{center}
\includegraphics[width=0.95\linewidth]{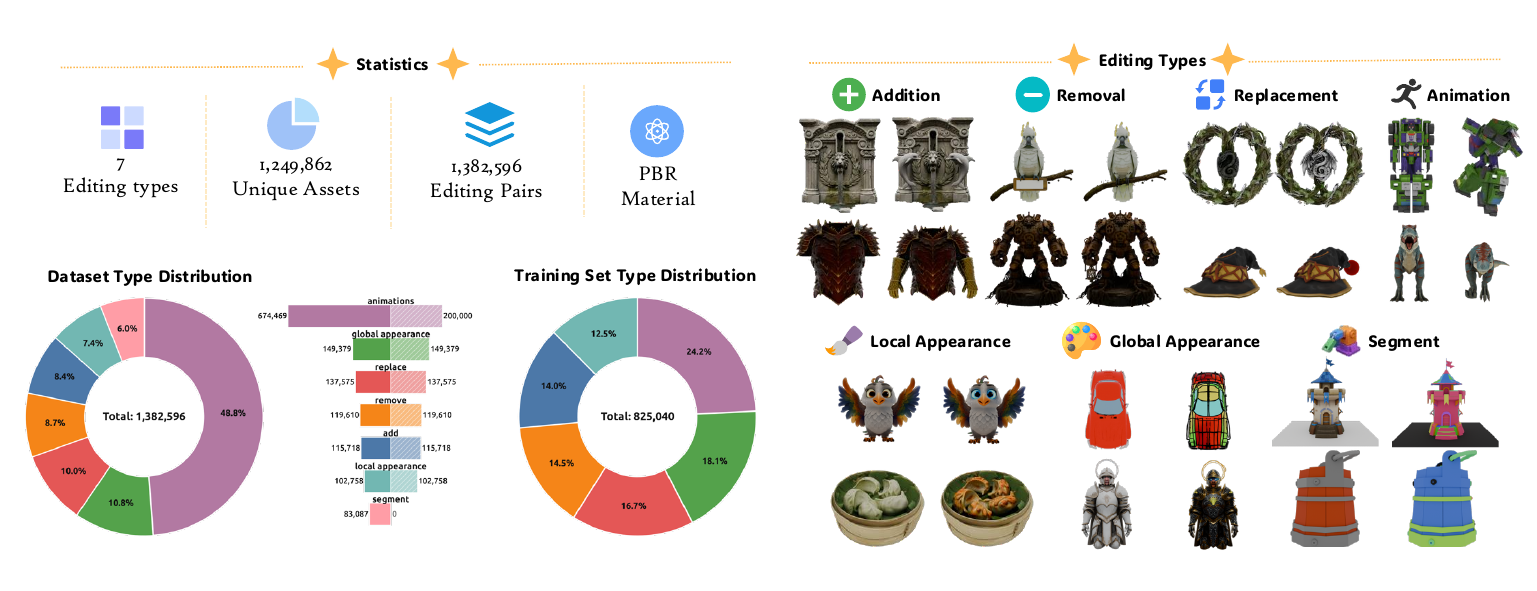}
\end{center}
\vspace{-0.1in}
\caption{Statistics of Alchemy3D-1M dataset. Alchemy3D-1M includes 1.25M unique assets and 1.38 M editing pairs, across 7 different editing types.}
\vspace{-0.1in}
\label{fig:dataset}
\end{figure}

As shown in Table~\ref{tab:dataset_comparison}, existing datasets for 3D asset editing remain limited in both scale and editing types. Among them, Steer3D~\citep{steer3d} independently reconstructs the before and after assets from corresponding image pairs, which can lead to poor identity preservation. Nano3D-100K and 3DEditVerse~\citep{nano3d, 3deditformer} focus primarily on structural edits, such as addition, removal, and replacement. PxForm, in contrast, curates editing pairs from labeled 3D part datasets~\citep{partverse}, limiting the diversity and complexity of the assets. In contrast, Alchemy3D-1M, to the best of our knowledge, is the first million-scale dataset for 3D asset editing.

The construction of Alchemy3D-1M is divided into three distinct categories. For structural editing data which cover addition, removal, replacement, as well as local and global appearance modifications, the dataset generation follows a structured three-stage pipeline consisting of \emph{preparation}, \emph{construction}, and \emph{filtering}. 
(1) In the data \emph{preparation} stage, source images are rendered from public 3D datasets~\citep{objaversexl,3dfuture,ShapeNet,abo,texverse,hssd} or generated using a text-to-image model~\citep{flux2}. The paired 3D source assets are re-generated with TRELLIS.2~\citep{trellis2} according to the source images, where the sampling trajectories are stored for subsequent operations.
Then, we use Qwen3-VL~\citep{Qwen3-VL} to write an editing text instruction for these source images and introduce FLUX.2-Dev-Turbo to generate target edited images.
(2) During data \emph{construction}, for removal and addition types, we detect~\citep{rexomni} and segment~\citep{sam3} 2D edit-region masks based on the text instruction and the rendered views from reconstructed assets. These masks are then lifted into 3D using SegViGen~\citep{segvigen}. For replacement type, the 3d editing mask is instead estimated by training-free editing method~\citep{flowedit}. Guided by these 3D masks, trajectory-aware 3D inpainting modifies the target region while maintaining the original diffusion trajectories in unedited areas to produce the edited 3D model. 
(3) Finally, in the \emph{filtering} stage, the generated 3D and image editing pairs are checked and recaptioned with Qwen3.6-27B~\citep{qwen3.6-27b}, where the low quality assets are discarded.

For animation data, we sample pairs of frames from motion sequences in existing animation datasets~\citep{objaversexl,texverse,deformingthings4d,mixamo} and articulation datasets~\citep{artiverse}. For articulated assets without explicit motion sequences, we simulate motions with a physics simulator~\citep{sapien}. We then compute inter-frame feature similarities with DINOv3~\citep{dinov3} and discard pairs without noticeable motion.

For segmentation data, we collect assets from 3D part segmentation datasets~\citep{partnet,PartNeXT,partversexl} and apply a custom color assignment algorithm to colorize individual parts. Rendered images of these colorized assets then serve as visual instructions to guide the segmentation.
The detailed construction pipeline is provided in Appendix~\ref{subsec:app_dataset}.

As shown in Table~\ref{tab:dataset_comparison} and Fig.~\ref{fig:dataset}, Alchemy3D-1M contains \textbf{1,249,862 unique 3D assets} and \textbf{1,382,596 annotated editing pairs} spanning \textbf{seven editing categories}: addition, removal, replacement, local appearance, global appearance, animation, and segmentation. For the training set, we downsample the animation pairs to 200k to balance the distribution across editing types and reserve the segmentation pairs exclusively for downstream evaluation. This yields a final training set of \textbf{825,040 editing pairs} for unified training. Compared to existing 3D editing datasets, Alchemy3D-1M is substantially larger, more diverse, and covers a broader spectrum of editing tasks, serving as a valuable asset for the 3D vision community.

\section{Alchemy3D Model}
\label{sec:model}

\subsection{Architecture.}

3D asset editing is inherently a generative task: one text or image instruction can yield multiple valid edited assets. 
While current editing approaches~\citep{3deditformer,partflow,anchorflow,steer3d} train generative models to modify 3D assets, they treat the source asset strictly as external guidance via an auxiliary branch such as ControlNet~\citep{controlnet}, which may miss crucial cross-feature interactions between the source and target assets.

\begin{figure}[t]
\centering
\includegraphics[width=0.95\linewidth]{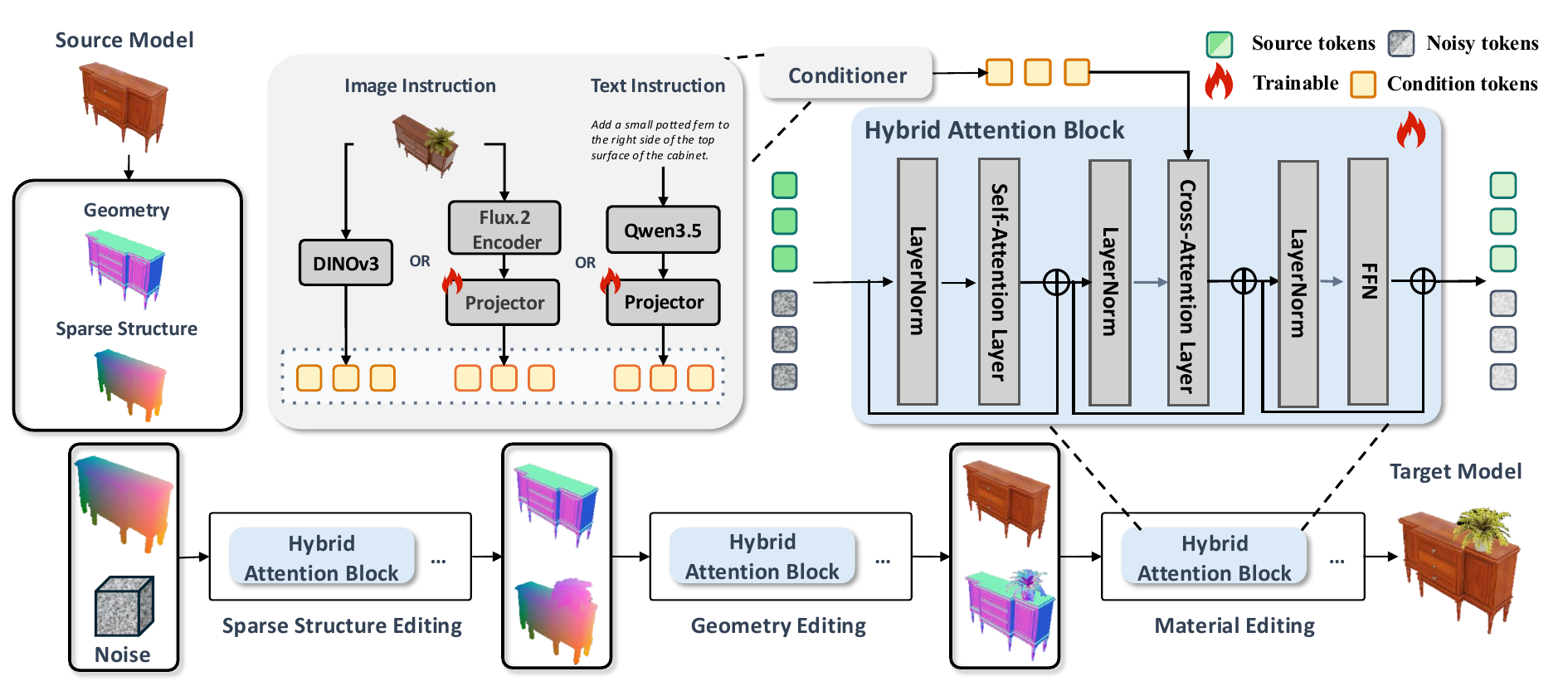}
\caption{Illustration of Alchemy3D model. Following Trellis.2, we design a three-stage editing pipeline that sequentially edits sparse structure (voxel occupancy), geometry (fine-grained surface), and material (PBR attributes). Each stage comprises hybrid attention blocks that denoise noisy target tokens by attending to source tokens from the source asset alongside condition tokens from image/text instructions. For clarity, we display decoded outputs from denoised tokens at each stage.}
\vspace{-0.2in}
\label{fig:model}
\end{figure}

As illustrated in Fig.~\ref{fig:model}, rather than introducing an auxiliary branch to inject source features, we directly incorporate tokens encoded from both the source asset and the noisy target state into the hybrid attention blocks, enabling in-depth feature interaction. Following Trellis.2~\citep{trellis2}, we organize these attention blocks into a three-stage flow transformer that sequentially edits sparse structure (voxel occupancy), geometry (fine-grained surface), and materials (PBR attributes).

The requested edit is specified by an external condition, which can be either a text instruction or an edited image instruction. To support diverse usage scenarios, we adopt multiple encoders to process different conditional signals. For \emph{text-conditioned} editing, we encode the text instruction with Qwen3.5-2B~\citep{qwen3.5} followed by a lightweight trainable projector, and the resulting features provide semantic guidance for specific editing operations.
For \emph{image-conditioned} editing, we use DINOv3~\citep{dinov3} to extract the condition tokens. Since DINO focuses on extracting semantic features instead of low-level vision features, we optionally trained a separate texture editing model with FLUX.2 encoder~\citep{flux2} followed by a lightweight projector for the material editing stage, which can help produce better color and texture fidelity. Comparison between DINOv3 and FLUX.2 encoder is presented in Sec.~\ref{sec:exp_results}.


\subsection{Training Objectives.}

\textbf{Flow Matching.}
During training, a pair of source and target assets is sampled and encoded into the corresponding latent representations, which can be sparse voxel, geometry, or material latents depending on the stage. The target latent $X_0$ is then interpolated with Gaussian noise according to $X_t=(1-t)X_0+tX_1$, and concatenated token-wise with the source latent $X'$. The generative flow model $v_\theta$ predicts the velocity field from the concatenated input tokens $[X'\ X_t]$, conditioned on the image or text instruction $c$ and the sampled timestep $t$. We then train the model with the standard optimal transport flow-matching objective~\citep{flowmatching}:
\begin{equation}
\label{eq:flow_matching}
\begin{aligned}
    &\mathcal{L}_{\text{FM}} = \mathbb{E}_{t, X_t \sim p_t}\left[||(X_1-X_0) - v_\theta(X', X_t, c,t)||_2^2\right],
\end{aligned}
\end{equation}

\textbf{Few-Step Distillation.}
While effective, concatenating source and target tokens roughly doubles the sequence length and consequently increases the computational cost of attention. We recover this efficiency through a two-stage distillation procedure inspired by few-step video distillation~\citep{anyflow}. In the first stage, we train a continuous-step flow map following MeanFlow~\citep{meanflow}, with the objective:
\begin{equation}
\label{eq:meanflow}
\begin{aligned}
    \mathcal{L}_{\text{MF}}(\theta)
    &=
    \mathbb{E}
    \left[
    \left\|
    u_\theta(X', X_t, c,r,t)
    -
    \operatorname{sg}(u_{\text{tgt}})
    \right\|_2^2
    \right], \\
    u_{\text{tgt}}
    &=
    v(X_t,t)
    -
    (t-r)
    \frac{d u_\theta(X',X_t, c,r,t)}{dt},
\end{aligned}
\end{equation}
where the flow map $u_\theta$, which differs slightly from the pretrained flow model, predicts the target flow conditioned on an additional timestep $r$. Here, $\operatorname{sg}(\cdot)$ denotes the stop-gradient operation, and $v$ denotes the pretrained flow model defined in Eq.~\ref{eq:flow_matching}. Following the transition model~\citep{transitionmodel}, we approximate the time derivative as
\begin{equation}
\frac{d u_\theta(X',X_t,c,r,t)}{dt}
\approx
\frac{
u_\theta(X',X_{t+\Delta t},c,r,t+\Delta t)
-
u_\theta(X',X_{t-\Delta t},c,r,t-\Delta t)
}{
2\Delta t
}.
\end{equation}
Starting from this flow map, we then perform on-policy distribution matching distillation following DMD~\citep{dmd,dmd2}. Given an initial Gaussian noise $X_1\sim\mathcal{N}(0,I)$, the student model first approximates a clean sample through its inference trajectory
$\hat{X_0}=f_\theta(X_1)$, which is then re-noised at a sampled timestep $t$ as
$X_t=(1-t)\hat{X_0}+t\epsilon$, where $\epsilon\sim\mathcal{N}(0,I)$. The DMD gradient is given by
\begin{equation}
\nabla_\theta \mathcal{L}_{\text{DMD}}
=
-\mathbb{E}_{t,X_1}
\left[
\left(
s_{\text{real}}(X_t,t)
-
s_{\text{fake}}(X_t,t)
\right)
\frac{\partial f_\theta(X_1)}{\partial\theta}
\right],
\end{equation}
where $s_{\text{real}}$ and $s_{\text{fake}}$ denote the score functions of the pretrained teacher and the student-generated distributions, respectively. We replace the adversarial loss in DMD2~\cite{dmd2} with the MeanFlow objective in Eq.~\ref{eq:meanflow}. Specifically, we optimize
\begin{equation}
\mathcal{L}_{\text{on-policy}}
=
\mathcal{L}_{\text{DMD}}
+
\mathcal{L}_{\text{MF}}.
\end{equation}
Throughout the few-step distillation process, only the attached LoRA modules are trainable.

\section{GEdit3D-Bench}
\label{sec:benchmark}
Existing benchmarks for 3D asset editing have several fundamental limitations. Edit3D-Bench, Eval3DEdit, and TANGOEdit~\citep{voxhammer,anchorflow,tango} each collect only about 100 assets from existing 3D datasets~\citep{objaversexl,gso,partobjaversetiny}, which is too small for a comprehensive evaluation. Another line of work, including Steer3D, Nano3D, 3DEditFormer, and PartFlow~\citep{steer3d,nano3d,3deditformer,partflow}, builds test sets by splitting the training data of the corresponding method, which risks leakage and overfitting to that specific distribution. The protocols are also narrow. The first line of works evaluates only CLIP or DINO similarity between rendered views and the input image. The second provides a low-quality ground-truth asset and measures alignment to it, which overlooks the quality of that reference and the one-to-many nature of generative editing. \textbf{GEdit3D-Bench} instead provides large-scale, open-world data and evaluates complementary aspects of editing quality.

\textbf{Construction.}Each sample contains a source asset $\mathcal{A}$, source and target captions $\mathcal{C}_{\mathrm{src}}$ and $\mathcal{C}_{\mathrm{tgt}}$, a source rendered image $\mathcal{I}_{\mathrm{src}}$, an editing instruction $\mathcal{I}$, and a target edited image $\mathcal{I}_{\mathrm{tgt}}$.
We first assemble synthetic assets generated with Hunyuan3D V3.1 and recently released Sketchfab assets from diverse categories. Then, we caption multi-view renderings, generate instructions for six editing types, and edit one source view to produce the visual target. Finally, automated checks and human review will be introduced to remove inconsistent or low-quality samples and generate target captions for text-based evaluation. 
More details can be found in Appendix~\ref{subsec:app_benchmark}.

\textbf{Metrics.}Given an edited asset $\mathcal{A}'$, we evaluate its multi-view renderings without treating a generated 3D target as ground truth. \textbf{View Quality} averages aesthetic-predictor-v2.5, MANIQA~\citep{maniqa}, and MUSIQ~\citep{musiq} scores across views~\citep{pyiqa}. 
\textbf{Reference Alignment} comprises image alignment $A_I$ and text alignment $A_T$, instantiated with complementary encoders including EVA-CLIP~\citep{evaclip}, SigLIP~\citep{siglip}, BLIP~\citep{blip}, DINOv3~\citep{dinov3}, and Uni3D~\citep{uni3d}, where applicable.
Following common practice in image and video evaluation~\citep{point2insert,goku,genaibench,imgedit}, we employ \textbf{MLLM (Multimodal Large Language Model) as judge}: \emph{Success Rate (SR)} is a binary measure of whether the requested edit succeeds, while \emph{Instruction Following (IF)}, \emph{Identity Preservation (IP)}, and \emph{Visual Quality (VQ)} are scored from 1 to 100.

\section{Experiments}
\label{sec:experiments}
\subsection{Evaluation Setting.}
We evaluate five variants of our model: (1) \textbf{Alchemy3D}, our default image-conditioned editor based on the DINOv3 encoder; (2) \textbf{Alchemy3D-Flux}, which uses the FLUX.2 image encoder to improve color and texture fidelity; (3) \textbf{Alchemy3D-Instruct}, which uses Qwen3.5-2B~\citep{qwen3.5} to support text-instruction input; (4) \textbf{Alchemy3D-Turbo}, a few-step image-conditioned editor distilled from Alchemy3D; and (5) \textbf{Alchemy3D-Segment}, which is fine-tuned from Alchemy3D for multi-view-guided 3D part segmentation.

For baselines and benchmarks, We compare image-conditioned Alchemy3D with Nano3D~\citep{nano3d}, 3DEditFormer~\citep{3deditformer}, and PartFlow~\citep{partflow}. 
For instruction-conditioned editing, we compare with Steer3D~\citep{steer3d}, 
In this work, we adopt GEdit3D-Bench (Sec.~\ref{sec:benchmark}) as our primary evaluation benchmark. For a more comprehensive comparison, we additionally evaluate on existing datasets, including Eval3DEdit~\citep{anchorflow}, 3DEditVerse~\citep{3deditformer}, Nano3D-100K~\citep{nano3d}, and the Edit3D-Bench introduced by VoxHammer~\citep{voxhammer}. Since Nano3D does not release an official test split, we randomly sample 1.5K pairs from its data for evaluation. For text-conditioned editors, we conduct comparisons on GEdit3D-Bench and on the Edit3D-Bench introduced by Steer3D.


Unless otherwise stated, MLLM scores use Gemini-3.8-Flash.\footnote{\url{https://deepmind.google/models/model-cards/gemini-3-8-flash/}} On GEdit3D-Bench, a random 400-sample subset is used for MLLM scoring and the user study. Alchemy3D and Alchemy3D-Instruct are sampled with 12 steps and classifier-free guidance on sparse structure and shape. Alchemy3D-Turbo uses 3 steps, with guidance fused during training.
Please refer to Sec.~\ref{subsubsec:exp_details} for more details.
\subsection{Experimental Results}
\label{sec:exp_results}
\begin{figure}[t]
    \centering
    \includegraphics[width=\linewidth]{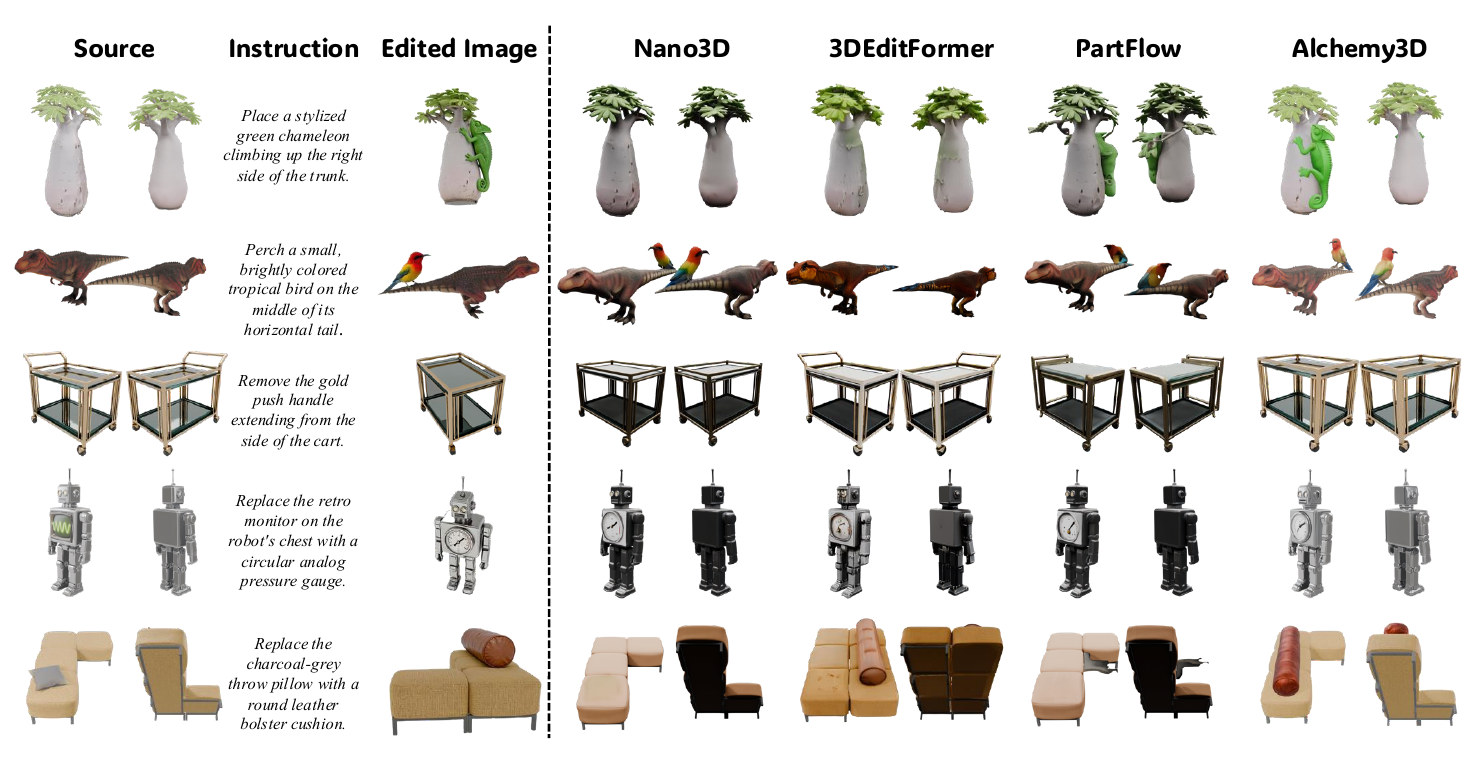}
    \vspace{-0.2in}
    \caption{Qualitative comparison between Alchemy3D and baselines. We show only addition, removal, and replacement, the editing types supported by all baselines.}
    \label{fig:qualitative_main}
    \vspace{-0.1in}
\end{figure}

\begin{figure}[h]
\begin{center}
\includegraphics[width=\linewidth]{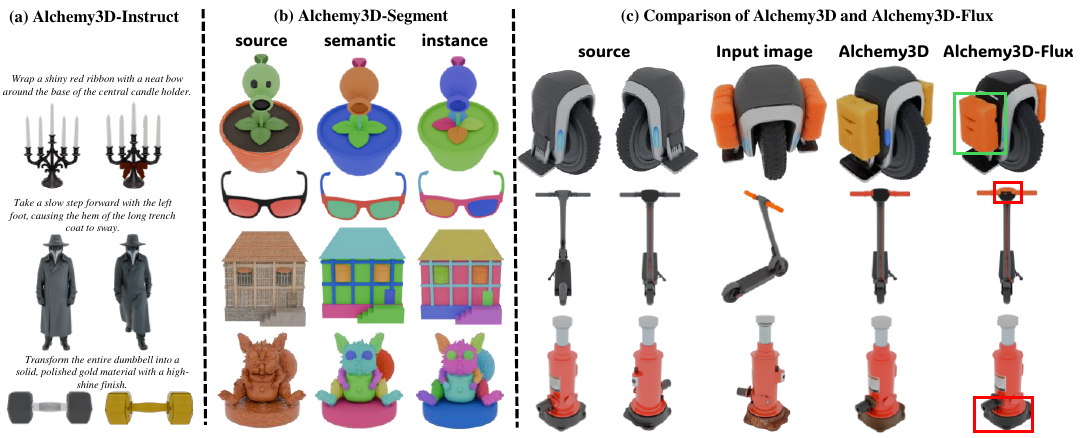}
\end{center}
\vspace{-0.1in}
\caption{(a) Examples from Alchemy3D-Instruct. (b) Segmentation results from Alchemy3D-Segment, where the segmentation criterion (e.g., semantic or instance) is controlled by the image instruction. (c) Comparison between Alchemy3D and Alchemy3D-Flux.}
\label{fig:qualitative_supp}
\end{figure}


\begin{table*}[!h]
    \centering
    \caption{Quantitative results on GEdit3D-Bench for addition, removal, and replacement. Per-type results are reported in Table~\ref{tab:gedit3d_full}. Human preference scores do not sum to 100\% because participants could also select a ``tie'' option when neither result was clearly better. Sampling time is measured on a single NVIDIA H200 GPU and averaged over 10 examples.}
    \vspace{-0.1in}
    \label{tab:results_main}
    \small
    \setlength{\tabcolsep}{1.5pt}
    \renewcommand{\arraystretch}{0.9}
    \begin{tabular}{l@{\hspace{2pt}}cc@{\hspace{2pt}}ccc@{\hspace{2pt}}cccc@{\hspace{2pt}}c@{\hspace{2pt}}c}
        \toprule
        & \multicolumn{2}{c}{\textbf{View Qual.}}
        & \multicolumn{3}{c}{\textbf{Ref. Alignment}}
        & \multicolumn{4}{c}{\textbf{MLLM}}
        & \textbf{Human}
        & \textbf{Time} \\
        \cmidrule(lr){2-3}
        \cmidrule(lr){4-6}
        \cmidrule(lr){7-10}
        \cmidrule(lr){11-11}
        \cmidrule(lr){12-12}
        \textbf{Method}
        & Aes. $\uparrow$
        & MUSIQ $\uparrow$
        & $A_I^{\mathrm{CLIP}}$ $\uparrow$
        & $A_T^{\mathrm{CLIP}}$ $\uparrow$
        & $A_I^{\mathrm{DINO}}$ $\uparrow$
        & SR $\uparrow$
        & VQ $\uparrow$
        & IF $\uparrow$
        & IP $\uparrow$
        & \textbf{Pref.}
        & \textbf{(s)} $\downarrow$ \\
        \midrule
        Nano3D
        & 4.40 & \underline{72.87}
        & 81.59 & 65.10 & 69.15
        & 51.14\% & \underline{70.16} & 53.84 & 72.09
        & 5.1\%
        & 8.629 \\
        3DEditFormer
        & 4.34 & 70.87
        & 81.83 & 63.19 & 68.07
        & 58.90\% & 66.31 & 59.79 & 63.39
        & 5.3\%
        & 3.262 \\
        PartFlow
        & 4.31 & 70.96
        & 81.25 & 62.24 & 67.78
        & 55.50\% & 65.27 & 56.20 & 66.60
        & 6.5\%
        & 7.184 \\
        \midrule
        \textbf{Alchemy3D-Turbo}
        & \textbf{4.58} & \textbf{72.94}
        & \underline{82.36} & \textbf{68.31} & \underline{72.88}
        & \underline{88.18\%} & 69.91 & \underline{70.71} & \underline{73.22}
        & --
        & \textbf{1.231} \\
        \textbf{Alchemy3D}
        & \underline{4.43} & 72.36
        & \textbf{84.91} & \underline{67.47} & \textbf{73.18}
        & \textbf{89.55\%} & \textbf{72.74} & \textbf{71.38} & \textbf{76.19}
        & \textbf{70.7\%}
        & 5.202 \\
        \bottomrule
    \end{tabular}
    \vspace{-0.2in}
\end{table*}

\begin{table*}[h]
    \centering
    \caption{Quantitative results on Eval3DEdit. Since Eval3DEdit originally only evaluates CLIP similarities, we additionally report aesthetic score for view quality and SR for MLLM rating.}
    \vspace{-0.1in}
    \label{tab:results_eval3dedit}
    \small
    \setlength{\tabcolsep}{2.5pt}
    \renewcommand{\arraystretch}{1.05}
    \begin{tabular}{lcccc cccc cccc}
        \toprule
        & \multicolumn{4}{c}{\textbf{Add / Remove / Replace}}
        & \multicolumn{4}{c}{\textbf{Action}}
        & \multicolumn{4}{c}{\textbf{Style}} \\
        \cmidrule(lr){2-5}
        \cmidrule(lr){6-9}
        \cmidrule(lr){10-13}
        \textbf{Method}
        & Aes. & $A_{I}$ & $A_T$ & SR
        & Aes. & $A_{I}$ & $A_T$ & SR
        & Aes. & $A_{I}$ & $A_T$ & SR \\
        \midrule
        Nano3D
        & \underline{4.60} & 82.31 & 49.43 & 40.00\%
        & -- & -- & -- & --
        & -- & -- & -- & -- \\
        3DEditFormer
        & 4.55 & 81.78 & 49.24 & 70.97\%
        & 4.06 & 75.80 & 44.45 & 20.00\%
        & -- & -- & -- & -- \\
        PartFlow
        & 4.51 & 81.59 & 49.32 & 67.86\%
        & -- & -- & -- & --
        & 4.58 & 78.10 & 50.68 & 40.00\% \\
        \midrule
        \textbf{Alchemy3D-Turbo}
        & \textbf{4.64} & \underline{82.67} & \textbf{52.56} & \textbf{90.00\%}
        & \textbf{4.33} & \underline{79.96} & \textbf{51.46} & \textbf{90.00\%}
        & \underline{4.61} & \underline{81.15} & \underline{52.51} & \textbf{100.00\%} \\
        \textbf{Alchemy3D}
        & \underline{4.60} & \textbf{83.77} & \underline{52.10} & \underline{83.87\%}
        & \underline{4.29} & \textbf{80.84} & \underline{51.15} & \underline{80.00\%}
        & \textbf{4.64} & \textbf{82.40} & \textbf{54.41} & \underline{90.00\%} \\
        \bottomrule
    \end{tabular}
    \vspace{-0.1in}
\end{table*}

\textbf{Image-conditioned editing.}
As demonstrated in Tables~\ref{tab:results_main} and~\ref{tab:results_eval3dedit}, \textbf{Alchemy3D} and \textbf{Alchemy3D-Turbo} outperform existing methods across both alignment and MLLM scores, with human preference for our methods significantly exceeding that of the baselines. More details on the human user study are provided in Sec.~\ref{subsubsec:exp_details}. Qualitative comparisons in Fig.~\ref{fig:qualitative_main} indicate our methods faithfully execute the requested changes while well preserving the identity of the source asset.
In terms of sampling efficiency, Alchemy3D-Turbo requires 1.231 seconds, making it faster than all baseline methods. Although PartFlow uses a lower-complexity ControlNet-style architecture, it samples with 50 steps by default and therefore incurs a substantial inference cost.
The remaining per-type results are reported in Table~\ref{tab:eval3dedit_full} and Table~\ref{tab:gedit3d_full}. As a side comparison, we also report image alignment scores on 3DEditVerse, the Edit3D-Bench introduced by VoxHammer, and Nano3D-100K in Table~\ref{tab:results_others}. Alchemy3D and Alchemy3D-Turbo still outperform the baselines.

\textbf{Instruction-conditioned editing.}
In this section, we compare text-conditioned editing between Alchemy3D-Instruct and Steer3D~\citep{steer3d}. Since Steer3D~\citep{steer3d} provides neither text captions nor image instructions, we evaluate editing performance using the aesthetic score for visual quality, together with the MLLM-based metrics SR, IF, and IP described in Sec.~\ref{sec:benchmark}.
As shown in Table~\ref{tab:results_instruct}, Alchemy3D-Instruct, trained across all editing types, outperforms Steer3D's type-specific models on both GEdit3D-Bench and Steer3D's Edit3D-Bench. Examples of Alchemy3D-Instruct are shown in Fig.~\ref{fig:qualitative_supp}(a).
Interestingly, the text-conditioned model underperforms the image-conditioned one, likely due to the ambiguity of text instructions. Motivated by this observation, we explore an agentic text-conditioned editing pipeline that automatically selects a 2D view, edits it, and then invokes the image-conditioned Alchemy3D. A detailed analysis is presented in Sec.~\ref{subsubsec:app_instruct_results}.

\textbf{Ablation on Encoder.}
As shown in Fig.~\ref{fig:qualitative_supp}(c), replacing DINOv3 with the FLUX.2 encoder improves color and style fidelity, while reducing local editing quality. Though the standard performances of the two models are close (Table~\ref{tab:results_flux}), Alchemy3D achieves much better performance when adapting to part segmentation with limited data (Table~\ref{tab:segmentation_results}). We therefore retain Alchemy3D as our default foundation and release Alchemy3D-Flux for appearance-critical use cases.

\textbf{Downstream part segmentation.}
To validate whether the priors acquired through unified editing transfer to downstream applications, we adapt Alchemy3D using LoRA~\citep{lora} to create \textbf{Alchemy3D-Segment}, a model designed for 3D part segmentation guided by 1 to 8 segmented views.
Results on PartObjaverse-Tiny~\citep{partobjaversetiny} (Table~\ref{tab:segmentation_results}) show that \textbf{Alchemy3D-Segment} substantially outperforms prior baselines and initialization variants trained from scratch or from TRELLIS.2. These improvements demonstrate that unified 3D editing serves as a powerful pretext task for downstream part understanding.
Interestingly, training Alchemy3D-Segment with variable 1-to-8 view inputs yields better 1-view evaluation performance than training on single views alone (Alchemy3D LoRA+1-view), highlighting the value of multi-view data augmentation.

\begin{table}[t]
    \centering
    \small
    \setlength{\tabcolsep}{3pt}
    \renewcommand{\arraystretch}{1.12}
\captionsetup{
    font=normalsize,
    labelfont=normalfont,
    labelsep=colon,
    justification=raggedright,
    singlelinecheck=false,
    skip=6pt,
    position=top
}
    \edef\AlchemyTableBase{\number\value{table}}
    \newcommand{\AlchemyCaption}[3]{%
        \setcounter{table}{\numexpr\AlchemyTableBase+#1\relax}%
        \caption{#2}\label{#3}%
    }

    \begin{minipage}[t]{0.48\linewidth}
        \vspace{0pt}

        \AlchemyCaption{0}{
            Comparison on other benchmark data.
        }{tab:results_others}

        \resizebox{\linewidth}{!}{%
            \begin{tabular}{@{}lcccc@{}}
                \toprule
                \textbf{Method}
                & $A_I^{\mathrm{CLIP}}$  $\uparrow$
                & $A_I^{\mathrm{SigLIP}}$ $\uparrow$
                & $A_I^{\mathrm{DINO}}$  $\uparrow$
                & $A_I^{\mathrm{Uni3D}}$  $\uparrow$ \\
                \midrule

                \rowcolor{gray!12}[0pt][0pt]
                \multicolumn{5}{@{}l@{}}{\textit{3DEditVerse}} \\
                3DEditFormer
                & 69.18 & 79.18 & 64.60 & \underline{36.76} \\
                PartFlow
                & 68.08 & 78.60 & 63.74 & 36.46 \\
                \textbf{Alchemy3D-Turbo}
                & \underline{72.10} & \underline{82.05}
                & \underline{69.44} & 36.36 \\
                \textbf{Alchemy3D}
                & \textbf{73.10} & \textbf{82.18}
                & \textbf{72.20} & \textbf{37.41} \\
                \midrule

                \rowcolor{gray!12}[0pt][0pt]
                \multicolumn{5}{@{}l@{}}{
                    \textit{Edit3D-Bench (VoxHammer)}
                } \\
                3DEditFormer
                & 76.91 & 84.99 & 66.90 & 38.33 \\
                PartFlow
                & 77.55 & 85.95 & 69.02 & \underline{38.93} \\
                \textbf{Alchemy3D-Turbo}
                & \underline{79.90} & \underline{87.28}
                & \underline{72.91} & 38.25 \\
                \textbf{Alchemy3D}
                & \textbf{80.91} & \textbf{88.44}
                & \textbf{75.04} & \textbf{39.06} \\
                \midrule

                \rowcolor{gray!12}[0pt][0pt]
                \multicolumn{5}{@{}l@{}}{\textit{Nano3D-100K}} \\
                3DEditFormer
                & 66.30 & 78.06 & 60.38 & 36.57 \\
                PartFlow
                & 65.42 & 78.61 & 61.18 & 36.20 \\
                \textbf{Alchemy3D-Turbo}
                & \underline{69.86} & \underline{80.81}
                & \underline{66.58} & \underline{37.04} \\
                \textbf{Alchemy3D}
                & \textbf{72.22} & \textbf{82.37}
                & \textbf{71.11} & \textbf{38.06} \\
                \bottomrule
            \end{tabular}%
        }

        \par\vspace{10pt}

        \AlchemyCaption{2}{
            Quantitative comparison between Steer3D and Alchemy3D-Instruct on GEdit3D-Bench and Edit3D-Bench (Steer3D).
        }{tab:results_instruct}

        \resizebox{\linewidth}{!}{%
            \begin{tabular}{@{}lcccc@{}}
                \toprule
                \textbf{Method}
                & \textbf{Aes.}
                & \textbf{SR}
                & \textbf{IF}
                & \textbf{IP} \\
                \midrule

                \rowcolor{gray!12}[0pt][0pt]
                \multicolumn{5}{@{}l@{}}{\textit{GEdit3D-Bench}} \\
                \textbf{Steer3D}
                & 3.84 & 12.37\% & 38.55 & 35.28 \\
                \textbf{Alchemy3D-Instruct}
                & \textbf{4.60} & \textbf{45.67\%}
                & \textbf{51.81} & \textbf{72.56} \\
                \midrule

                \rowcolor{gray!12}[0pt][0pt]
                \multicolumn{5}{@{}l@{}}{
                    \textit{Edit3D-Bench (Steer3D)}
                } \\
                \textbf{Steer3D}
                & 3.78 & 33.73\% & 38.69 & 53.94 \\
                \textbf{Alchemy3D-Instruct}
                & \textbf{3.94} & \textbf{49.60\%}
                & \textbf{67.81} & \textbf{71.76} \\
                \bottomrule
            \end{tabular}%
        }
    \end{minipage}%
    \hfill
    \begin{minipage}[t]{0.48\linewidth}
        \vspace{0pt}

        \AlchemyCaption{1}{
            Comparison between Alchemy3D
            and Alchemy3D-Flux on GEdit3D-Bench. 
        }{tab:results_flux}

        \resizebox{\linewidth}{!}{%
            \begin{tabular}{@{}lcccc@{}}
                \toprule
                \textbf{Model}
                & Aes. $\uparrow$
                & $A_I^{\mathrm{CLIP}}$ $\uparrow$
                & $A_T^{\mathrm{CLIP}}$ $\uparrow$
                & $A_I^{\mathrm{DINO}}$ $\uparrow$ \\
                \midrule
                Alchemy3D-Flux
                & \textbf{4.420} & 83.43
                & \textbf{66.36} & 72.26 \\
                Alchemy3D
                & 4.419 & \textbf{83.53}
                & \textbf{66.36} & \textbf{72.52} \\
                \bottomrule
            \end{tabular}%
        }

        \par\vspace{4pt}

        \AlchemyCaption{3}{
            Quantitative results for semantic part
            segmentation on PartObjaverse-Tiny. Trial experiments compare different strategies for training single-view conditioned segmentation.
        }{tab:segmentation_results}

        \begingroup
        \footnotesize
        \renewcommand{\arraystretch}{1.02}

        \begin{tabular*}{\linewidth}{
            @{\extracolsep{\fill}}lr@{}
        }
            \toprule
            \textbf{Method} & \textbf{mIoU} $\uparrow$ \\
            \midrule

            \rowcolor{gray!12}[0pt][0pt]
            \multicolumn{2}{@{}l@{}}{\textit{Trial Experiments}} \\
            From Scratch (1-view)              & 28.54 \\
            From TRELLIS.2 (1-view)            & 58.68 \\
            From Alchemy3D (1-view)            & 60.08 \\
            From Alchemy3D-Flux (LoRA+1-view) & 52.55 \\
            From Alchemy3D (LoRA+1-view)      & 61.10 \\
            \midrule

            \rowcolor{gray!12}[0pt][0pt]
            \multicolumn{2}{@{}l@{}}{\textit{Baselines}} \\
            Find3D    & 19.35 \\
            PartField & 52.01 \\
            P3-SAM    & 46.58 \\
            SegViGen  & 53.49 \\
            
            \midrule

            \rowcolor{gray!12}[0pt][0pt]
            \multicolumn{2}{@{}l@{}}{\textit{Ours}} \\
            \textbf{Alchemy3D-Segment(1 to 8-views)} & \\
            \hspace{1.5em}{\scriptsize (1-view)} & 63.10 \\
            \hspace{1.5em}{\scriptsize (2-view)} & 65.23 \\
            \hspace{1.5em}{\scriptsize (4-view)} & 66.98 \\
            \hspace{1.5em}{\scriptsize (8-view)} & \textbf{68.70} \\
            \bottomrule
        \end{tabular*}
        \endgroup
    \end{minipage}
    \vspace{-0.2in}
\end{table}

\section{Conclusion}

In this work, we present \textbf{Alchemy3D}, a unified framework spanning scalable data construction, model architecture, and benchmark evaluation for versatile 3D editing. We construct \textbf{Alchemy3D-1M}, a dataset of 1.38 million editing pairs across seven editing types, and introduce an editing architecture that fuses tokens from source and edited target 3D representations via self-attention, with text/image instructions injected via cross-attention. Together with a few-step distillation procedure, this architecture yields a series of \textbf{Alchemy3D} model variants tailored to different editing use cases. To enable fair, comprehensive evaluation, we also introduce \textbf{GEdit3D-Bench}, a large-scale, multi-dimensional benchmark entirely independent of our training data. Extensive experiments show that our models achieve significant improvements over existing 3D editing methods, with the distilled variant achieving a $4\times$ acceleration while maintaining superior performance.

\clearpage
\section{AI use statement}
In this work, we used generative AI tools to edit the manuscript for readability and to assist with writing and debugging code. We have reviewed all AI-assisted work. Two authors checked the revised text, and the AI-assisted code was tested for correctness. We take responsibility for the final content of this work, including text, claims, or artifacts produced with the aid of generative AI.

\bibliography{main}

@inproceedings{ddpm,
  author={Jonathan Ho and Ajay Jain and Pieter Abbeel},
  title={Denoising Diffusion Probabilistic Models},
  booktitle={Advances in Neural Information Processing Systems 33},
  year={2020}
}

@inproceedings{ncsn,
  author={Yang Song and Stefano Ermon},
  title={Generative Modeling by Estimating Gradients of the Data Distribution},
  booktitle={Advances in Neural Information Processing Systems},
  year={2019},
}

@inproceedings{score-sde,
  author={Yang Song and Jascha Sohl{-}Dickstein and Diederik P. Kingma and Abhishek Kumar and Stefano Ermon and Ben Poole},
  title={Score-Based Generative Modeling through Stochastic Differential Equations},
  booktitle={9th International Conference on Learning Representations},
  year={2021},
}

@inproceedings{flowmatching,
  author={Yaron Lipman and Ricky T. Q. Chen and Heli Ben{-}Hamu and Maximilian Nickel and Matthew Le},
  title={Flow Matching for Generative Modeling},
  booktitle={The Eleventh International Conference on Learning Representations},
  year={2023},
}

@article{tailor3d,
  author={Zhangyang Qi and Yunhan Yang and Mengchen Zhang and Long Xing and Xiaoyang Wu and Tong Wu and Dahua Lin and Xihui Liu and Jiaqi Wang and Hengshuang Zhao},
  title={Tailor3D: Customized 3D Assets Editing and Generation with Dual-Side Images},
  journal      = {CoRR},
  year         = {2024},
  eprinttype   = {arXiv},
  eprint       = {2407.06191},
}

@article{editcast3d,
  title={EditCast3D: Single-Frame-Guided 3D Editing with Video Propagation and View Selection},
  author={Qu, Huaizhi and Zhang, Ruichen and Luo, Shuqing and Qi, Luchao and Zhang, Zhihao and Liu, Xiaoming and Sengupta, Roni and Chen, Tianlong},
  journal={arXiv preprint arXiv:2510.13652},
  year={2025}
}

@inproceedings{edit360,
  title={Edit360: 2d image edits to 3d assets from any angle},
  author={Huang, Junchao and Hu, Xinting and Shi, Shaoshuai and Tian, Zhuotao and Jiang, Li},
  booktitle={Proceedings of the IEEE/CVF International Conference on Computer Vision},
  year={2025}
}

@article{pro3deditor,
  title={Pro3d-editor: A progressive-views perspective for consistent and precise 3d editing},
  author={Zheng, Yang and Huang, Mengqi and Chen, Nan and Mao, Zhendong},
  journal={Advances in Neural Information Processing Systems},
  year={2025}
}

@inproceedings{instructnerf2nerf,
  title={Instruct-nerf2nerf: Editing 3d scenes with instructions},
  author={Haque, Ayaan and Tancik, Matthew and Efros, Alexei A and Holynski, Aleksander and Kanazawa, Angjoo},
  booktitle={Proceedings of the IEEE/CVF international conference on computer vision},
  year={2023}
}

@inproceedings{voxe,
  title={Vox-e: Text-guided voxel editing of 3d objects},
  author={Sella, Etai and Fiebelman, Gal and Hedman, Peter and Averbuch-Elor, Hadar},
  booktitle={Proceedings of the IEEE/CVF international conference on computer vision},
  year={2023}
}

@article{gsedit,
  title={Gsedit: Efficient text-guided editing of 3d objects via gaussian splatting},
  author={Palandra, Francesco and Sanchietti, Andrea and Baieri, Daniele and Rodola, Emanuele},
  journal={arXiv preprint arXiv:2403.05154},
  year={2024}
}

@inproceedings{focaldreamer,
  title={Focaldreamer: Text-driven 3d editing via focal-fusion assembly},
  author={Li, Yuhan and Dou, Yishun and Shi, Yue and Lei, Yu and Chen, Xuanhong and Zhang, Yi and Zhou, Peng and Ni, Bingbing},
  booktitle={Proceedings of the AAAI conference on artificial intelligence},
  year={2024}
}

@article{steer3d,
  title={Feedforward 3D Editing via Text-Steerable Image-to-3D},
  author={Ma, Ziqi and Chen, Hongqiao and Yue, Yisong and Gkioxari, Georgia},
  journal={arXiv preprint arXiv:2512.13678},
  year={2025}
}

@article{3deditformer,
  title={Towards Scalable and Consistent 3D Editing},
  author={Xia, Ruihao and Tang, Yang and Zhou, Pan},
  journal={arXiv preprint arXiv:2510.02994},
  year={2025}
}

@article{partflow,
  title={Feedforward 3D Editing Learns from Semantic-Part Transformation},
  author={Weng, Jiawei and Zhang, Saining and Diao, Zhenxin and Li, Peishuo and Zhang, Henghaofan and Chen, Junhao and Zhao, Hao},
  journal={arXiv preprint arXiv:2605.27351},
  year={2026}
}

@inproceedings{voxhammer,
  title={Voxhammer: Training-free precise and coherent 3d editing in native 3d space},
  author={Li, Lin and Huang, Zehuan and Feng, Haoran and Zhuang, Gengxiong and Chen, Rui and Guo, Chunchao and Sheng, Lu},
  booktitle={2026 International Conference on 3D Vision (3DV)},
  year={2026},
}

@article{nano3d,
  title={NANO3D: A Training-Free Approach for Efficient 3D Editing Without Masks},
  author={Ye, Junliang and Xie, Shenghao and Zhao, Ruowen and Wang, Zhengyi and Yan, Hongyu and Zu, Wenqiang and Ma, Lei and Zhu, Jun},
  journal={arXiv preprint arXiv:2510.15019},
  year={2025}
}

@inproceedings{anchorflow,
  title={Anchorflow: Training-free 3d editing via latent anchor-aligned flows},
  author={Zhou, Zhenglin and Ma, Fan and Gui, Chengzhuo and Xia, Xiaobo and Fan, Hehe and Yang, Yi and Chua, Tat-Seng},
  booktitle={Proceedings of the IEEE/CVF Conference on Computer Vision and Pattern Recognition},
  year={2026}
}

@article{tango,
  title={TanGO: Training-Free 3D Editing via Tangent-Space Guidance and Optimization},
  author={Lim, Siwoo and Yoon, Sunjae and Koo, Gwanhyeong and Yun, Hyeonseo and Yoo, Chang D},
  journal={arXiv preprint arXiv:2607.14927},
  year={2026}
}

@inproceedings{trellis,
  author={Jianfeng Xiang and Zelong Lv and Sicheng Xu and Yu Deng and Ruicheng Wang and Bowen Zhang and Dong Chen and Xin Tong and Jiaolong Yang},
  title={Structured 3D Latents for Scalable and Versatile 3D Generation},
  booktitle={Proceedings of the IEEE/CVF Conference on Computer Vision and Pattern Recognition},
  year={2025},
}

@inproceedings{trellis2,
  title={Native and compact structured latents for 3d generation},
  author={Xiang, Jianfeng and Chen, Xiaoxue and Xu, Sicheng and Wang, Ruicheng and Lv, Zelong and Deng, Yu and Zhu, Hongyuan and Dong, Yue and Zhao, Hao and Yuan, Nicholas Jing and others},
  booktitle={Proceedings of the IEEE/CVF Conference on Computer Vision and Pattern Recognition},
  year={2026}
}

@article{hunyuan3d2.1,
  title={Hunyuan3d 2.1: From images to high-fidelity 3d assets with production-ready pbr material},
  author={Hunyuan3D, Team and Yang, Shuhui and Yang, Mingxin and Feng, Yifei and Huang, Xin and Zhang, Sheng and He, Zebin and Luo, Di and Liu, Haolin and Zhao, Yunfei and others},
  journal={arXiv preprint arXiv:2506.15442},
  year={2025}
}

@inproceedings{dreamfusion,
  author={Ben Poole and Ajay Jain and Jonathan T. Barron and Ben Mildenhall},
  title={DreamFusion: Text-to-3D using 2D Diffusion},
  booktitle={The Eleventh International Conference on Learning Representations},
  year={2023},
}

@inproceedings{pointdiffusion,
  title={Diffusion probabilistic models for 3d point cloud generation},
  author={Luo, Shitong and Hu, Wei},
  booktitle={Proceedings of the IEEE/CVF conference on computer vision and pattern recognition},
  year={2021}
}

@article{pointe,
  title={Point-e: A system for generating 3d point clouds from complex prompts},
  author={Nichol, Alex and Jun, Heewoo and Dhariwal, Prafulla and Mishkin, Pamela and Chen, Mark},
  journal={arXiv preprint arXiv:2212.08751},
  year={2022}
}

@inproceedings{diffrf,
  title={Diffrf: Rendering-guided 3d radiance field diffusion},
  author={M{\"u}ller, Norman and Siddiqui, Yawar and Porzi, Lorenzo and Bulo, Samuel Rota and Kontschieder, Peter and Nie{\ss}ner, Matthias},
  booktitle={Proceedings of the IEEE/CVF conference on computer vision and pattern recognition},
  year={2023}
}

@inproceedings{rodin,
  title={Rodin: A generative model for sculpting 3d digital avatars using diffusion},
  author={Wang, Tengfei and Zhang, Bo and Zhang, Ting and Gu, Shuyang and Bao, Jianmin and Baltrusaitis, Tadas and Shen, Jingjing and Chen, Dong and Wen, Fang and Chen, Qifeng and others},
  booktitle={Proceedings of the IEEE/CVF conference on computer vision and pattern recognition},
  year={2023}
}

@inproceedings{gvgen,
  title={Gvgen: Text-to-3d generation with volumetric representation},
  author={He, Xianglong and Chen, Junyi and Peng, Sida and Huang, Di and Li, Yangguang and Huang, Xiaoshui and Yuan, Chun and Ouyang, Wanli and He, Tong},
  booktitle={European Conference on Computer Vision},
  year={2024},
}

@article{nerf,
  title={Nerf: Representing scenes as neural radiance fields for view synthesis},
  author={Mildenhall, Ben and Srinivasan, Pratul P and Tancik, Matthew and Barron, Jonathan T and Ramamoorthi, Ravi and Ng, Ren},
  journal={Communications of the ACM},
  year={2021}
}

@article{3dgs,
    author = {Kerbl, Bernhard and Kopanas, Georgios and Leimkuehler, Thomas and Drettakis, George},
    title = {3D Gaussian Splatting for Real-Time Radiance Field Rendering},
    year = {2023},
    journal = {ACM Trans. Graph.},
}

@inproceedings{anyedit,
  title={Anyedit: Mastering unified high-quality image editing for any idea},
  author={Yu, Qifan and Chow, Wei and Yue, Zhongqi and Pan, Kaihang and Wu, Yang and Wan, Xiaoyang and Li, Juncheng and Tang, Siliang and Zhang, Hanwang and Zhuang, Yueting},
  booktitle={Proceedings of the Computer Vision and Pattern Recognition Conference},
  year={2025}
}

@article{hunyuanimage3,
  title={HunyuanImage 3.0 Technical Report},
  author={Cao, Siyu and Chen, Hangting and Chen, Peng and Cheng, Yiji and Cui, Yutao and Deng, Xinchi and Dong, Ying and Gong, Kipper and Gu, Tianpeng and Gu, Xiusen and others},
  journal={arXiv preprint arXiv:2509.23951},
  year={2025}
}

@article{rfinversion,
  title={Semantic image inversion and editing using rectified stochastic differential equations},
  author={Rout, Litu and Chen, Yujia and Ruiz, Nataniel and Caramanis, Constantine and Shakkottai, Sanjay and Chu, Wen-Sheng},
  journal={arXiv preprint arXiv:2410.10792},
  year={2024}
}

@inproceedings{rfsolver,
  title={Taming Rectified Flow for Inversion and Editing},
  author={Wang, Jiangshan and Pu, Junfu and Qi, Zhongang and Guo, Jiayi and Ma, Yue and Huang, Nisha and Chen, Yuxin and Li, Xiu and Shan, Ying},
  booktitle={International Conference on Machine Learning},
  year={2025},
}

@inproceedings{flowedit,
  title={Flowedit: Inversion-free text-based editing using pre-trained flow models},
  author={Kulikov, Vladimir and Kleiner, Matan and Huberman-Spiegelglas, Inbar and Michaeli, Tomer},
  booktitle={Proceedings of the IEEE/CVF International Conference on Computer Vision},
  year={2025}
}

@inproceedings{imgedit,
  author={Yang Ye and Xianyi He and Zongjian Li and Bin Lin and Shenghai Yuan and Zhiyuan Yan and Bohan Hou and Li Yuan},
  title={ImgEdit: {A} Unified Image Editing Dataset and Benchmark},
  booktitle={Advances in Neural Information Processing Systems 38: Annual Conference on Neural Information Processing Systems 2025, NeurIPS 2025, San Diego, CA, USA, December 2-7, 2025 / Mexico City, Mexico, November 30 - December 5, 2025},
  year={2025}
}

@article{genaibench,
  title={Genai arena: An open evaluation platform for generative models},
  author={Jiang, Dongfu and Ku, Max and Li, Tianle and Ni, Yuansheng and Sun, Shizhuo and Fan, Rongqi and Chen, Wenhu},
  journal={Advances in Neural Information Processing Systems},
  year={2024}
}

@inproceedings{controlnet,
  title={Adding conditional control to text-to-image diffusion models},
  author={Zhang, Lvmin and Rao, Anyi and Agrawala, Maneesh},
  booktitle={2023 IEEE/CVF International Conference on Computer Vision (ICCV)},
  year={2023}
}

@article{point2insert,
  title={Point2Insert: Video Object Insertion via Sparse Point Guidance},
  author={Zhou, Yu and Yang, Xiaoyan and Zi, Bojia and Zhang, Lihan and Sun, Ruijie and Zheng, Weishi and Huang, Haibin and Zhang, Chi and Li, Xuelong},
  journal={arXiv preprint arXiv:2602.04167},
  year={2026}
}

@inproceedings{vbench,
  title={Vbench: Comprehensive benchmark suite for video generative models},
  author={Huang, Ziqi and He, Yinan and Yu, Jiashuo and Zhang, Fan and Si, Chenyang and Jiang, Yuming and Zhang, Yuanhan and Wu, Tianxing and Jin, Qingyang and Chanpaisit, Nattapol and others},
  booktitle={Proceedings of the IEEE/CVF Conference on Computer Vision and Pattern Recognition},
  year={2024}
}

@article{goku,
  title={Goku: A Million-Scale Universal Dataset and Benchmark for Instruction-Based Video Editing},
  author={Liang, Sen and Wang, Cong and Yu, Zhentao and Guan, Fengbin and Zhou, Zhengguang and Hu, Teng and Zhang, Youliang and Zhou, Yuan and Li, Xin and Lu, Qinglin and others},
  journal={arXiv preprint arXiv:2606.30599},
  year={2026}
}

@inproceedings{rexomni,
  title={Detect anything via next point prediction},
  author={Jiang, Qing and Huo, Junan and Chen, Xingyu and Xiong, Yuda and Zeng, Zhaoyang and Chen, Yihao and Ren, Tianhe and Yu, Junzhi and Zhang, Lei},
  booktitle={Proceedings of the IEEE/CVF Conference on Computer Vision and Pattern Recognition},
  year={2026}
}

@article{sam2,
  title={SAM 2: Segment Anything in Images and Videos},
  author={Ravi, Nikhila and Gabeur, Valentin and Hu, Yuan-Ting and Hu, Ronghang and Ryali, Chaitanya and Ma, Tengyu and Khedr, Haitham and R{\"a}dle, Roman and Rolland, Chloe and Gustafson, Laura and Mintun, Eric and Pan, Junting and Alwala, Kalyan Vasudev and Carion, Nicolas and Wu, Chao-Yuan and Girshick, Ross and Doll{\'a}r, Piotr and Feichtenhofer, Christoph},
  journal={arXiv preprint arXiv:2408.00714},
  url={https://arxiv.org/abs/2408.00714},
  year={2024}
}

@article{sam3,
  author={Nicolas Carion and Laura Gustafson and Yuan{-}Ting Hu and Shoubhik Debnath and Ronghang Hu and Didac Suris and Chaitanya Ryali and Kalyan Vasudev Alwala and Haitham Khedr and Andrew Huang and Jie Lei and Tengyu Ma and Baishan Guo and Arpit Kalla and Markus Marks and Joseph Greer and Meng Wang and Peize Sun and Roman R{\"{a}}dle and Triantafyllos Afouras and Effrosyni Mavroudi and Katherine Xu and Tsung{-}Han Wu and Yu Zhou and Liliane Momeni and Rishi Hazra and Shuangrui Ding and Sagar Vaze and Francois Porcher and Feng Li and Siyuan Li and Aishwarya Kamath and Ho Kei Cheng and Piotr Doll{\'{a}}r and Nikhila Ravi and Kate Saenko and Pengchuan Zhang and Christoph Feichtenhofer},
  title={{SAM} 3: Segment Anything with Concepts},
  year={2025},
  eprinttype={arXiv},
  eprint={2511.16719}
}

@article{segvigen,
  title={SegviGen: Repurposing 3D Generative Model for Part Segmentation},
  author={Li, Lin and Feng, Haoran and Huang, Zehuan and Chen, Haohua and Nie, Wenbo and Hou, Shaohua and Fan, Keqing and Hu, Pan and Wang, Sheng and Li, Buyu and others},
  journal={arXiv preprint arXiv:2603.16869},
  year={2026}
}

@article{Qwen3-VL,
      title={Qwen3-VL Technical Report}, 
      author={Shuai Bai and Yuxuan Cai and Ruizhe Chen and Keqin Chen and Xionghui Chen and Zesen Cheng and Lianghao Deng and Wei Ding and Chang Gao and Chunjiang Ge and Wenbin Ge and Zhifang Guo and Qidong Huang and Jie Huang and Fei Huang and Binyuan Hui and Shutong Jiang and Zhaohai Li and Mingsheng Li and Mei Li and Kaixin Li and Zicheng Lin and Junyang Lin and Xuejing Liu and Jiawei Liu and Chenglong Liu and Yang Liu and Dayiheng Liu and Shixuan Liu and Dunjie Lu and Ruilin Luo and Chenxu Lv and Rui Men and Lingchen Meng and Xuancheng Ren and Xingzhang Ren and Sibo Song and Yuchong Sun and Jun Tang and Jianhong Tu and Jianqiang Wan and Peng Wang and Pengfei Wang and Qiuyue Wang and Yuxuan Wang and Tianbao Xie and Yiheng Xu and Haiyang Xu and Jin Xu and Zhibo Yang and Mingkun Yang and Jianxin Yang and An Yang and Bowen Yu and Fei Zhang and Hang Zhang and Xi Zhang and Bo Zheng and Humen Zhong and Jingren Zhou and Fan Zhou and Jing Zhou and Yuanzhi Zhu and Ke Zhu},
	  journal={arXiv preprint arXiv:2511.21631},
      year={2025}
}

@misc{qwen3.6-27b,
    titl ={{Qwen3.6-27B}: Flagship-Level Coding in a {27B} Dense Model},
    author={{Qwen Team}},
    year={2026},
    url={https://qwen.ai/blog?id=qwen3.6-27b}
}

@misc{qwen3.5,
    title  = {{Qwen3.5}: Towards Native Multimodal Agents},
    author = {{Qwen Team}},
    year   = {2026},
    month  = {February},
    url    = {https://qwen.ai/blog?id=qwen3.5}
}

@misc{flux2,
    author={Black Forest Labs},
    title={{FLUX.2: Frontier Visual Intelligence}},
    year={2025},
    howpublished={\url{https://bfl.ai/blog/flux-2}},
}

@article{cosmos3,
  title={Cosmos 3: Omnimodal world models for physical ai},
  author={Agarwal, Niket and Ali, Arslan and Allen, Jon and Antolini, Martin and Aubame, Adeline and Azzolini, Alisson and Bai, Junjie and Bala, Maciej and Balaji, Yogesh and Bapst, Josh and others},
  journal={arXiv preprint arXiv:2606.02800},
  year={2026}
}

@article{texverse,
  title={Texverse: A universe of 3d objects with high-resolution textures},
  author={Zhang, Yibo and Zhang, Li and Ma, Rui and Cao, Nan},
  journal={arXiv preprint arXiv:2508.10868},
  year={2025}
}

@misc{mixamo,
    author={Adobe Systems Inc.},
    title={{Mixamo.}},
    year={2025},
    howpublished={\url{https://www.mixamo.com/}},
}

@article{objaversexl,
  title={Objaverse-xl: A universe of 10m+ 3d objects},
  author={Deitke, Matt and Liu, Ruoshi and Wallingford, Matthew and Ngo, Huong and Michel, Oscar and Kusupati, Aditya and Fan, Alan and Laforte, Christian and Voleti, Vikram and Gadre, Samir Yitzhak and others},
  journal={Advances in Neural Information Processing Systems},
  year={2023}
}

@inproceedings{abo,
  title={Abo: Dataset and benchmarks for real-world 3d object understanding},
  author={Collins, Jasmine and Goel, Shubham and Deng, Kenan and Luthra, Achleshwar and Xu, Leon and Gundogdu, Erhan and Zhang, Xi and Vicente, Tomas F Yago and Dideriksen, Thomas and Arora, Himanshu and others},
  booktitle={Proceedings of the IEEE/CVF conference on computer vision and pattern recognition},
  year={2022}
}

@article{3dfuture,
  title={3d-future: 3d furniture shape with texture},
  author={Fu, Huan and Jia, Rongfei and Gao, Lin and Gong, Mingming and Zhao, Binqiang and Maybank, Steve and Tao, Dacheng},
  journal={International Journal of Computer Vision},
  year={2021},
}

@inproceedings{hssd,
  title={Habitat synthetic scenes dataset (hssd-200): An analysis of 3d scene scale and realism tradeoffs for objectgoal navigation},
  author={Khanna, Mukul and Mao, Yongsen and Jiang, Hanxiao and Haresh, Sanjay and Shacklett, Brennan and Batra, Dhruv and Clegg, Alexander and Undersander, Eric and Chang, Angel X and Savva, Manolis},
  booktitle={Proceedings of the IEEE/CVF Conference on Computer Vision and Pattern Recognition},
  year={2024}
}

@inproceedings{deformingthings4d,
  title={4dcomplete: Non-rigid motion estimation beyond the observable surface},
  author={Li, Yang and Takehara, Hikari and Taketomi, Takafumi and Zheng, Bo and Nie{\ss}ner, Matthias},
  booktitle={Proceedings of the IEEE/CVF International Conference on Computer Vision},
  year={2021}
}

@inproceedings{partverse,
  title={From one to more: Contextual part latents for 3d generation},
  author={Dong, Shaocong and Ding, Lihe and Chen, Xiao and Li, Yaokun and Wang, Yuxin and Wang, Yucheng and Wang, Qi and Kim, Jaehyeok and Gao, Chenjian and Huang, Zhanpeng and others},
  booktitle={2025 IEEE/CVF International Conference on Computer Vision (ICCV)},
  year={2025}
}

@article{partversexl,
  title={FullPart: Generating each 3D Part at Full Resolution},
  author={Ding, Lihe and Dong, Shaocong and Li, Yaokun and Gao, Chenjian and Chen, Xiao and Han, Rui and Kuang, Yihao and Zhang, Hong and Huang, Bo and Huang, Zhanpeng and others},
  journal={arXiv preprint arXiv:2510.26140},
  year={2025}
}

@inproceedings{sapien,
    author={Xiang, Fanbo and Qin, Yuzhe and Mo, Kaichun and Xia, Yikuan and Zhu, Hao and Liu, Fangchen and Liu, Minghua and Jiang, Hanxiao and Yuan, Yifu and Wang, He and Yi, Li and Chang, Angel X. and Guibas, Leonidas J. and Su, Hao},
    title={{SAPIEN}: A SimulAted Part-based Interactive ENvironment},
    booktitle={The IEEE Conference on Computer Vision and Pattern Recognition (CVPR)},
    year={2020}
}

@inproceedings{partnet,
  title={Partnet: A large-scale benchmark for fine-grained and hierarchical part-level 3d object understanding},
  author={Mo, Kaichun and Zhu, Shilin and Chang, Angel X and Yi, Li and Tripathi, Subarna and Guibas, Leonidas J and Su, Hao},
  booktitle={Proceedings of the IEEE/CVF conference on computer vision and pattern recognition},
  year={2019}
}

@inproceedings{artiverse,
  title={Artiverse: A Diverse and Physically Grounded Dataset for Articulated Objects},
  author={Iliash, Denys and Liu, Jiayi and Fokin, Egor and Wu, Qirui and Mahdavi-Amiri, Ali and Savva, Manolis and Chang, Angel X.},
  booktitle={Proceedings of the IEEE/CVF Conference on Computer Vision and Pattern Recognition (CVPR)},
  year={2026}
}

@inproceedings{gso,
  title={Google scanned objects: A high-quality dataset of 3d scanned household items},
  author={Downs, Laura and Francis, Anthony and Koenig, Nate and Kinman, Brandon and Hickman, Ryan and Reymann, Krista and McHugh, Thomas B and Vanhoucke, Vincent},
  booktitle={2022 International Conference on Robotics and Automation (ICRA)},
  year={2022},
}

@article{partobjaversetiny,
  title={Sampart3d: Segment any part in 3d objects},
  author={Yang, Yunhan and Huang, Yukun and Guo, Yuan-Chen and Lu, Liangjun and Wu, Xiaoyang and Lam, Edmund Y and Cao, Yan-Pei and Liu, Xihui},
  journal={arXiv preprint arXiv:2411.07184},
  year={2024}
}

@inproceedings{PartNeXT,
  author={Penghao Wang and Yiyang He and Xin Lv and Yukai Zhou and Lan Xu and Jingyi Yu and Jiayuan Gu},
  title={PartNeXt: {A} Next-Generation Dataset for Fine-Grained and Hierarchical 3D Part Understanding},
  booktitle={Advances in Neural Information Processing Systems 38: Annual Conference on Neural Information Processing Systems 2025, NeurIPS 2025, San Diego,CA, USA, December 2-7, 2025 / Mexico City, Mexico, November 30 - December5, 2025},
  year={2025}
}

@inproceedings{diffusion4d,
  author={Hanwen Liang and Yuyang Yin and Dejia Xu and Hanxue Liang and Zhangyang Wang and Konstantinos N. Plataniotis and Yao Zhao and Yunchao Wei},
  title={Diffusion4D: Fast Spatial-temporal Consistent 4D generation via Video Diffusion Models},
  booktitle={Advances in Neural Information Processing Systems 37: Annual Conference on Neural Information Processing Systems 2024, NeurIPS 2024, Vancouver, BC, Canada, December 10 - 15, 2024},
  year={2024}
}

@article{ShapeNet,
  author={Angel X. Chang and Thomas A. Funkhouser and Leonidas J. Guibas and Pat Hanrahan and Qi{-}Xing Huang and Zimo Li and Silvio Savarese and Manolis Savva and Shuran Song and Hao Su and Jianxiong Xiao and Li Yi and Fisher Yu},
  title={ShapeNet: An Information-Rich 3D Model Repository},
  year={2015},
  eprinttype={arXiv},
}

@inproceedings{clip,
  title={Learning transferable visual models from natural language supervision},
  author={Radford, Alec and Kim, Jong Wook and Hallacy, Chris and Ramesh, Aditya and Goh, Gabriel and Agarwal, Sandhini and Sastry, Girish and Askell, Amanda and Mishkin, Pamela and Clark, Jack and others},
  booktitle={International conference on machine learning},
  year={2021}
}

@article{evaclip,
  title={Eva-clip: Improved training techniques for clip at scale},
  author={Sun, Quan and Fang, Yuxin and Wu, Ledell and Wang, Xinlong and Cao, Yue},
  journal={arXiv preprint arXiv:2303.15389},
  year={2023}
}

@inproceedings{siglip,
  title={Sigmoid loss for language image pre-training},
  author={Zhai, Xiaohua and Mustafa, Basil and Kolesnikov, Alexander and Beyer, Lucas},
  booktitle={Proceedings of the IEEE/CVF international conference on computer vision},
  year={2023}
}

@inproceedings{blip,
  title={Blip: Bootstrapping language-image pre-training for unified vision-language understanding and generation},
  author={Li, Junnan and Li, Dongxu and Xiong, Caiming and Hoi, Steven},
  booktitle={International conference on machine learning},
  year={2022},
  organization={PMLR}
}

@article{dinov3,
  title={Dinov3},
  author={Sim{\'e}oni, Oriane and Vo, Huy V and Seitzer, Maximilian and Baldassarre, Federico and Oquab, Maxime and Jose, Cijo and Khalidov, Vasil and Szafraniec, Marc and Yi, Seungeun and Ramamonjisoa, Micha{\"e}l and others},
  journal={arXiv preprint arXiv:2508.10104},
  year={2025}
}

@inproceedings{uni3d,
  title={Uni3d: Exploring unified 3d representation at scale},
  author={Zhou, Junsheng and Wang, Jinsheng and Ma, Baorui and Liu, Yu-Shen and Huang, Tiejun and Wang, Xinlong},
  booktitle={International Conference on Learning Representations},
  year={2024}
}

@misc{pyiqa,
  title={{IQA-PyTorch}: PyTorch Toolbox for Image Quality Assessment},
  author={Chaofeng Chen and Jiadi Mo},
  year={2022},
  howpublished = "[Online]. Available: \url{https://github.com/chaofengc/IQA-PyTorch}"
}

@inproceedings{maniqa,
  title={MANIQA: Multi-dimension Attention Network for No-Reference Image Quality Assessment},
  author={Yang, Sidi and Wu, Tianhe and Shi, Shuwei and Lao, Shanshan and Gong, Yuan and Cao, Mingdeng and Wang, Jiahao and Yang, Yujiu},
  booktitle={Proceedings of the IEEE/CVF Conference on Computer Vision and Pattern Recognition},
  year={2022}
}

@misc{musiq,
      title={MUSIQ: Multi-scale Image Quality Transformer}, 
      author={Junjie Ke and Qifei Wang and Yilin Wang and Peyman Milanfar and Feng Yang},
      year={2021},
      eprint={2108.05997},
      archivePrefix={arXiv},
      primaryClass={cs.CV},
      url={https://arxiv.org/abs/2108.05997}, 
}

@inproceedings{dmd,
  author={Tianwei Yin and Micha{\"{e}}l Gharbi and Richard Zhang and Eli Shechtman and Fr{\'{e}}do Durand and William T. Freeman and Taesung Park},
  title={One-Step Diffusion with Distribution Matching Distillation},
  booktitle={{IEEE/CVF} Conference on Computer Vision and Pattern Recognition, {CVPR} 2024, Seattle, WA, USA, June 16-22, 2024},
  year={2024}
}

@inproceedings{dmd2,
  author={Tianwei Yin and Micha{\"{e}}l Gharbi and Taesung Park and Richard Zhang and Eli Shechtman and Fr{\'{e}}do Durand and Bill Freeman},
  title={Improved Distribution Matching Distillation for Fast Image Synthesis},
  booktitle={Advances in Neural Information Processing Systems 37: Annual Conference on Neural Information Processing Systems 2024, NeurIPS 2024, Vancouver, BC, Canada, December 10 - 15, 2024},
  year={2024}
}

@misc{anyflow,
    title={AnyFlow: Any-Step Video Diffusion Model with On-Policy Flow Map Distillation}, 
    author={Yuchao Gu and Guian Fang and Yuxin Jiang and Weijia Mao and Song Han and Han Cai and Mike Zheng Shou},
    year={2026},
    eprint={2605.13724},
    archivePrefix={arXiv}
}

@article{meanflow,
  title={Mean flows for one-step generative modeling},
  author={Geng, Zhengyang and Deng, Mingyang and Bai, Xingjian and Kolter, Zico and He, Kaiming},
  journal={Advances in Neural Information Processing Systems},
  year={2025}
}

@inproceedings{transitionmodel,
  title={Transition models: Rethinking the generative learning objective},
  author={Wang, Zidong and Zhang, Yiyuan and Yue, Xiaoyu and Yue, Xiangyu and Li, Yangguang and Ouyang, Wanli and Bai, Lei},
  booktitle={Proceedings of the IEEE/CVF Conference on Computer Vision and Pattern Recognition},
  year={2026}
}

@article{ssim,
  author={Zhou Wang and Alan C. Bovik and Hamid R. Sheikh and Eero P. Simoncelli},
  title={Image quality assessment: from error visibility to structural similarity},
  journal={{IEEE} Trans. Image Process.},
  year={2004},
}

@inproceedings{lpips,
  author={Richard Zhang and Phillip Isola and Alexei A. Efros and Eli Shechtman and Oliver Wang},
  title={The Unreasonable Effectiveness of Deep Features as a Perceptual Metric},
  booktitle={2018 {IEEE} Conference on Computer Vision and Pattern Recognition, {CVPR} 2018, Salt Lake City, UT, USA, June 18-22, 2018},
  year={2018}
}

@inproceedings{cd,
  author={Haoqiang Fan and Hao Su and Leonidas J. Guibas},
  title={A Point Set Generation Network for 3D Object Reconstruction from a Single Image},
  booktitle={2017 {IEEE} Conference on Computer Vision and Pattern Recognition, {CVPR} 2017, Honolulu, HI, USA, July 21-26, 2017},
  year={2017},
}

@inproceedings{nc,
  author={Georgia Gkioxari and Justin Johnson and Jitendra Malik},
  title={Mesh {R-CNN}},
  booktitle={2019 {IEEE/CVF} International Conference on Computer Vision, {ICCV} 2019, Seoul, Korea (South), October 27 - November 2, 2019},
  year={2019},
}

@article{f1,
  author={Arno Knapitsch and Jaesik Park and Qian{-}Yi Zhou and Vladlen Koltun},
  title={Tanks and temples: benchmarking large-scale scene reconstruction},
  journal={{ACM} Trans. Graph.},
  year={2017},
}

@article{moe,
  title={Adaptive mixtures of local experts},
  author={Jacobs, Robert A and Jordan, Michael I and Nowlan, Steven J and Hinton, Geoffrey E},
  journal={Neural computation},
  year={1991}
}

@inproceedings{lora,
  author={Edward J. Hu and Yelong Shen and Phillip Wallis and Zeyuan Allen{-}Zhu and Yuanzhi Li and Shean Wang and Lu Wang and Weizhu Chen},
  title ={LoRA: Low-Rank Adaptation of Large Language Models},
  booktitle={The Tenth International Conference on Learning Representations, {ICLR} 2022, Virtual Event, April 25-29, 2022},
  year={2022}
}

@article{vae,
  title={Auto-encoding variational bayes},
  author={Kingma, Diederik P and Welling, Max},
  journal={arXiv preprint arXiv:1312.6114},
  year={2013}
}
\bibliographystyle{iclr2027_conference}

\clearpage

\appendix
\section{Appendix}
\subsection{Model Training Details and Compute Report}
\label{subsec:app_model}


\subsubsection{Hyperparameters}
\paragraph{LoRA.} For Alchemy3D-Segment, we attach a LoRA module to the query, key, value, and output projections of each transformer block, with rank 96. For Alchemy3D-Turbo, in addition to these projection layers, we attach LoRA modules to the FFN, AdaLN modulation, input and output layers, and the two time embedders, with rank 256.

\paragraph{Classifier-free Guidance.} Both MeanFlow~\citep{meanflow} and DMD~\citep{dmd,dmd2} fuse classifier-free guidance into the model during training. We follow most of the guidance settings used when sampling the pretrained generator. We found, however, that a MeanFlow model with guidance disabled, as used in the original sampler, produced noisy textures. We therefore set the guidance strength to 3.0 when training the third-stage PBR editing model.

\subsubsection{Training Compute Report}
We report the core training configurations and compute resources in Tab.~\ref{tab:training_config} to facilitate reproduction.
\begin{table}[ht]
    \centering
    \caption{Training configurations of our models. Stage 1 refers to sparse-structure editing, stage 2 to Shape SLat editing, and stage 3 to PBR SLat editing.}
    \label{tab:training_config}
    \small
    \setlength{\tabcolsep}{0.5pt}
    \renewcommand{\arraystretch}{1.15}
    \begin{tabular}{l c l c c c c}
        \toprule
        \textbf{Model Name} &
        \textbf{Stage} &
        \textbf{Initialization} &
        \makecell{\textbf{Training}\\\textbf{Steps}} &
        \makecell{\textbf{Batch}\\\textbf{Size}} &
        \makecell{\textbf{\# GPUs}\\\textbf{(NVIDIA H200)}} &
        \makecell{\textbf{Training}\\\textbf{Time (hours)}} \\
        \midrule
        Alchemy3D & 1 & TRELLIS.2 & 62.5K & 192 & 16 & $\sim$83.33 \\
        Alchemy3D & 2 & TRELLIS.2 & 65K & 128 & 16 & $\sim$133.61 \\
        Alchemy3D & 3 & TRELLIS.2 & 65K & 128 & 16 & $\sim$162.50 \\
        Alchemy3D-Flux & 3 & TRELLIS.2-Flux & 50K & 128 & 16 & $\sim$125 \\
        Alchemy3D-Instruct & 1 & TRELLIS.2-Instruct & 75K & 192 & 16 & $\sim$105.83 \\
        Alchemy3D-Instruct & 2 & TRELLIS.2-Instruct & 40K & 128 & 16 & $\sim$100 \\
        Alchemy3D-Instruct & 3 & TRELLIS.2-Instruct & 60K & 96 & 16 & $\sim$100 \\
        Alchemy3D-Segment (Trial) & 3 & Alchemy3D & 10K & 64 & 8 & $\sim$10.83 \\
        Alchemy3D-Segment & 3 & Alchemy3D-Segment (Trial) & 20K & 64 & 8 & $\sim$28.89 \\
        Alchemy3D-MF & 1 & Alchemy3D & 15K & 288 & 24 & $\sim$74.44 \\
        Alchemy3D-MF & 2 & Alchemy3D & 15K & 192 & 24 & $\sim$65.83 \\
        Alchemy3D-MF & 3 & Alchemy3D & 10K & 192 & 24 & $\sim$55.83 \\
        Alchemy3D-Turbo & 1 & Alchemy3D-MF & 12K & 288 & 24 & $\sim$56.67 \\
        Alchemy3D-Turbo & 2 & Alchemy3D-MF & 12K & 96 & 24 & $\sim$37.78 \\
        Alchemy3D-Turbo & 3 & Alchemy3D-MF & 14K & 96 & 24 & $\sim$43.33 \\
        \bottomrule
    \end{tabular}
\end{table}

\subsection{Alchemy3D-1M Construction Pipeline}
\label{subsec:app_dataset}
In this section, we describe the construction pipeline of Alchemy3D-1M (Sec.~\ref{sec:dataset}). It has three stages: asset preparation, type-specific construction, and data filtering with condition generation.
\subsubsection{Stage 1: Asset Preparation}
\paragraph{Animation.}
For animation editing, we collect assets with temporal motion from multiple public datasets, including the animated subset of Objaverse-XL~\citep{objaversexl} filtered by Diffusion4D~\citep{diffusion4d}, TexVerse~\citep{texverse}, and DeformingThings4D~\citep{deformingthings4d}. To further diversify character-motion combinations, we randomly pair rigged characters from Mixamo~\citep{mixamo} with different motion sequences. We additionally incorporate articulated assets from ArtiVerse~\citep{artiverse} and PartNet-Mobility~\citep{partnet}.

\paragraph{Segmentation.}
For part segmentation, we collect assets with dense semantic part annotations from PartNet~\citep{partnet}, PartNeXT~\citep{PartNeXT}, and PartVerseXL~\citep{partversexl}. These annotations are directly used to construct the corresponding segmentation supervision.

\paragraph{Other editing types.}
For the remaining five edit categories, namely addition, removal, replacement, local appearance editing, and global appearance editing, we adopt a shared preparation pipeline. We first construct a large pool of source images by rendering assets from existing 3D datasets~\citep{objaversexl,3dfuture,ShapeNet,abo,texverse,hssd}, together with internally generated asset-centric images synthesized by a text-to-image foundation model~\citep{flux2}. Given each source image, Qwen3-VL~\citep{Qwen3-VL} generates edit instructions according to edit-type-specific prompting templates. The source image and generated instruction are then provided to FLUX.2-Dev-Turbo~\citep{flux2} to synthesize the corresponding edited target image. The generated source-target image pairs are subsequently verified using Qwen3-VL to remove semantically inconsistent or low-quality edits.

For each retained source image, we reconstruct the corresponding source 3D asset using TRELLIS.2~\citep{trellis2}. To ensure the quality of the reconstructed assets, we render multiple views and employ Qwen3-VL to assess their geometry and appearance, discarding assets with noticeable reconstruction artifacts or quality issues. For the accepted source assets, we additionally preserve the intermediate diffusion trajectories, which are later reused for trajectory-aware 3D inpainting to preserve the unedited regions during editing.

\subsubsection{Stage 2: Type-specific Construction Pipeline}
\paragraph{Animation.}
For assets with existing motion sequences, we uniformly sample keyframes at fixed temporal intervals and randomly pair different keyframes of the same asset to construct animation editing pairs. For articulated assets, we use SAPIEN~\citep{sapien} to simulate joint interactions for each annotated joint, obtain the resulting motion trajectories, and similarly sample and randomly pair keyframes from the same asset.

\paragraph{Segmentation.}
For each asset with $K$ annotated parts, we construct a $K$-color palette with well-separated hues. We first randomly sample the initial hue $h_0$ and generate the remaining hues using the golden-ratio increment:
\begin{equation}
\begin{aligned}
h_0 &\sim \mathcal{U}\left[0,1\right), \\
h_{i+1} &= (h_i+\phi)\ \text{mod}\ 1,
\end{aligned}
\end{equation}
where $\phi=\frac{\sqrt{5}-1}{2}\approx0.618$ is the golden-ratio increment. For each part, the saturation and value are independently sampled as $s_i\sim\mathcal{U}\left[0.7,0.9\right]$ and $v_i\sim\mathcal{U}\left[0.8,0.95\right]$. The resulting HSV colors are converted to RGB and assigned to mesh faces according to their annotated part labels, producing the corresponding part segmentation representation.

\paragraph{Addition/Removal.}
Addition and removal are constructed using a shared segment-and-remove pipeline. We independently construct image-editing vocabularies for the two editing types rather than deriving one exclusively by reversing the other, thereby avoiding the limited diversity and potential distributional bias introduced by relying on a single editing direction.

For each image pair, we reconstruct the asset in the state where the edited object is present using TRELLIS.2. For removal, this reconstructed asset directly serves as the source asset. For addition, it instead serves as the target asset, and the resulting pair is reversed after constructing its object-absent counterpart.

We render multiple views of the reconstructed asset and employ Rex-Omni~\citep{rexomni} to identify the object to be removed. SAM3~\citep{sam3} is subsequently used to obtain dense 2D segmentations. Among all rendered views, we select the view that jointly maximizes the visible area and the number of detected instances of the segmented object, thereby retaining as many target components as possible. The selected rendering is then converted into a two-color map, where the segmented object instances are colored white and the remaining objects are colored dark gray. Together with the corresponding 3D asset, this map is provided to SegViGen~\citep{segvigen} to recover the editable region in 3D. We further apply binary clustering to the predicted segmentation, assuming two clusters corresponding to the editable and unedited regions, and use the resulting cluster assignment to construct the final binary 3D mask.

Given the recovered 3D mask, we construct the object-absent counterpart through trajectory-aware 3D inpainting. Instead of re-noising the reconstructed asset during sampling, we directly reuse the diffusion trajectory recorded when generating the original asset. At each sampling step, the latent in the unedited region is replaced with the corresponding latent from the stored trajectory:
\begin{equation}
\begin{aligned}
    &\tilde{z}_t =
    x_t \odot (1-M) + z_t' \odot M, \\
    &z_{t-\Delta t}' =
    \tilde{z}_t - \Delta t\, v_\theta(\tilde{z}_t,t,c),
\end{aligned}
\end{equation}
where $\Delta t>0$ is the sampling interval, $M$ is the binary 3D mask with $M=1$ indicating the editable region, $x_t$ is the latent from the stored diffusion trajectory of the original asset at timestep $t$, and $z_t'$ is the current latent generated under the edited condition $c$. The mask is projected to the corresponding representations, and the same procedure is applied throughout the sparse-structure, Shape SLat, and PBR SLat generation stages. Consequently, the unedited regions remain explicitly constrained by the original generation trajectory throughout the reconstruction process.

\paragraph{Replacement.}
Unlike addition and removal, replacement edits simultaneously involve the removal of an existing object and the introduction of a new one, making the edited region more challenging to localize through direct segmentation. We therefore estimate the editable region by comparing the sparse structures before and after editing.

Specifically, we employ FlowEdit~\citep{flowedit} to approximately transform the source sparse structure conditioned on the edited image. The spatial differences between the resulting structure and the original source structure are used to estimate the editable region, from which we construct a binary 3D mask. Importantly, FlowEdit is used only for estimating the spatial extent of the edit rather than generating the final edited asset. The estimated mask is subsequently used in the trajectory-aware 3D inpainting procedure described above to generate the final geometry and appearance while preserving the unedited regions of the source asset.

\paragraph{Local Appearance.}
For local appearance editing, we adopt the same 3D mask construction procedure as in addition and removal to localize the edited region. We preserve the sparse structure of the source asset and perform mask-guided 3D inpainting only during the Shape SLat and PBR SLat generation stages. We do not hold the Shape SLat fixed, because appearance modifications, such as material or style changes, may also require adjustments to fine geometric details while preserving the overall structure of the asset.

\paragraph{Global Appearance.}
For global appearance editing, no spatial mask is required. We preserve the sparse structure of the source asset and regenerate the Shape SLat and PBR SLat conditioned on the edited image, allowing the appearance of the asset to be globally modified while maintaining its overall structure.

\subsubsection{Stage 3: Data Filtering and Condition Generation}
The preceding stages produce candidate editing pairs, which may still contain reconstruction artifacts, insufficient editing changes, or other quality issues. We therefore design edit-type-specific filtering procedures to remove undesirable samples and improve the overall quality of Alchemy3D-1M. For the retained samples, we further generate editing instructions and image-based conditions for subsequent model training.

\paragraph{Animation.}
To remove samples with insufficient motion changes, we render eight fixed circular views of both the source and target assets. For each corresponding view, we compute the cosine similarity between their DINOv3~\citep{dinov3} features:
\begin{equation}
    s_v = \operatorname{cos}\!\left(
    f_{\mathrm{DINO}}(R_v(X_s)),
    f_{\mathrm{DINO}}(R_v(X_t))
    \right), \quad v=1,\ldots,8.
\end{equation}
We retain a pair only if
\begin{equation}
    \min_{v=1,\ldots,8} s_v < \tau,
\end{equation}
where $\tau$ is a predefined similarity threshold. This filtering removes pairs for which the motion change is insufficiently apparent from all considered viewpoints. For each retained pair, the eight source and target renderings are concatenated into a grid and provided to Qwen3.6-27B~\citep{qwen3.6-27b} to generate editing instructions describing the motion changes. To increase linguistic diversity, we generate 3--5 instructions for each pair using different phrasings while preserving the same editing semantics. For image-conditioned editing, we additionally render eight random views of the target asset as image conditions.

\paragraph{Segmentation.}
For segmentation editing, we render eight random views of the part-colored assets using flat colors rather than conventional lighting. This rendering scheme is designed to mimic practical 2D segmentation-map conditions, where the regions predicted by a 2D segmentation model~\citep{sam2,sam3} can be represented as manually assigned flat colors. The resulting multi-view renderings are used as image conditions for segmentation editing.

\paragraph{Other editing types.}
For addition, removal, replacement, local appearance, and global appearance editing, we render multiple views of each editing pair and concatenate them into grid images for quality filtering and recaptioning with Qwen3.6-27B~\citep{qwen3.6-27b}. For addition, removal, replacement, and local appearance editing, we additionally render 16 random views of each target asset and use Qwen3.6-27B to determine whether the edited region is visible in each view. Views in which the edited region is not sufficiently visible are discarded. For global appearance editing, where the change is not spatially localized, we render eight random views instead.
\subsection{GEdit3D-Bench}
\label{subsec:app_benchmark}
\begin{figure}[t]
    \centering
    \includegraphics[width=\linewidth]{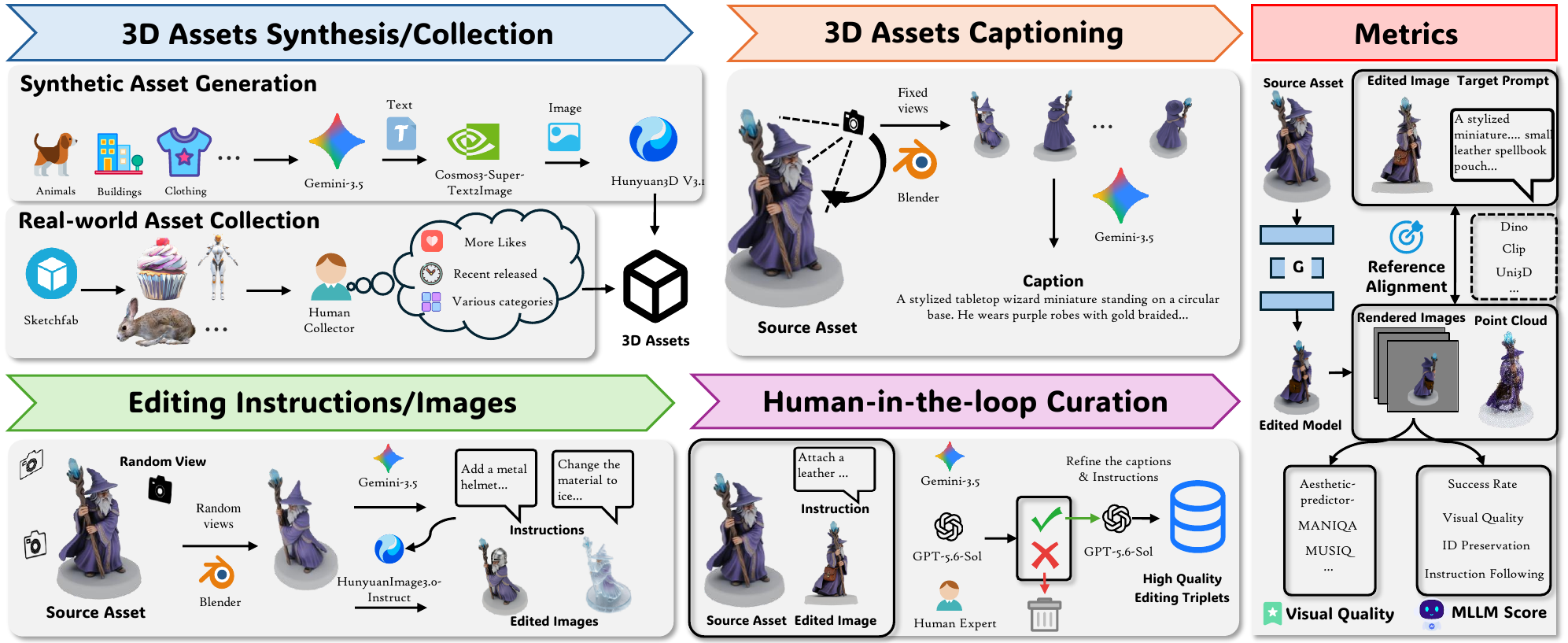}
    \caption{Overview of the construction pipeline and evaluation dimensions of GEdit3D-Bench.}
    \label{fig:bench}
\end{figure}

\subsubsection{Limitations of Existing Benchmarks}
The quality of a benchmark and its evaluation protocol can strongly influence the development of a research field. A meaningful benchmark should provide a fair, reliable, and realistic assessment of model capabilities while remaining aligned with practical applications. However, existing benchmarks for 3D asset editing still suffer from several fundamental limitations.

\paragraph{Limited Scale and Diversity.}
Existing benchmarks typically evaluate models on a relatively small number of manually curated assets. For example, Edit3D-Bench~\citep{voxhammer} selects only 100 assets from GSO~\citep{gso} and PartObjectVerse-Tiny~\citep{partobjaversetiny}. Similarly, Eval3DEdit~\citep{anchorflow} and TANGOEdit~\citep{tango} curate approximately 100 assets from datasets including Objaverse-XL~\citep{objaversexl} and GSO~\citep{gso}. While these benchmarks provide valuable initial evaluations, their limited scale restricts the coverage of asset categories, appearances, geometric structures, and editing scenarios, making it difficult to assess the robustness of 3D editing models in diverse real-world settings.

\paragraph{Low-Quality Ground Truth and In-distribution Evaluation.}
Another line of work~\citep{steer3d,nano3d,3deditformer,partflow} constructs evaluation sets by splitting the training data of the corresponding method into training and test subsets. Models are then evaluated by measuring the similarity between generated outputs and the provided edited 3D assets. This protocol has several limitations. First, existing 3D editing datasets contain annotation and generation errors, including Alchemy3D-1M (Sec.~\ref{sec:dataset}). Treating generated or reconstructed assets as ground truth can therefore pass those artifacts into the score. Second, 3D asset editing is a conditional generation task with multiple valid solutions, so similarity to a single reference is an incomplete measure of editing quality. Third, a random split of the same distribution used for training measures in-distribution fit. A model may score well by matching dataset-specific patterns without generalizing to unseen assets and edits.

\begin{figure}[h]
\begin{center}
\includegraphics[width=\linewidth]{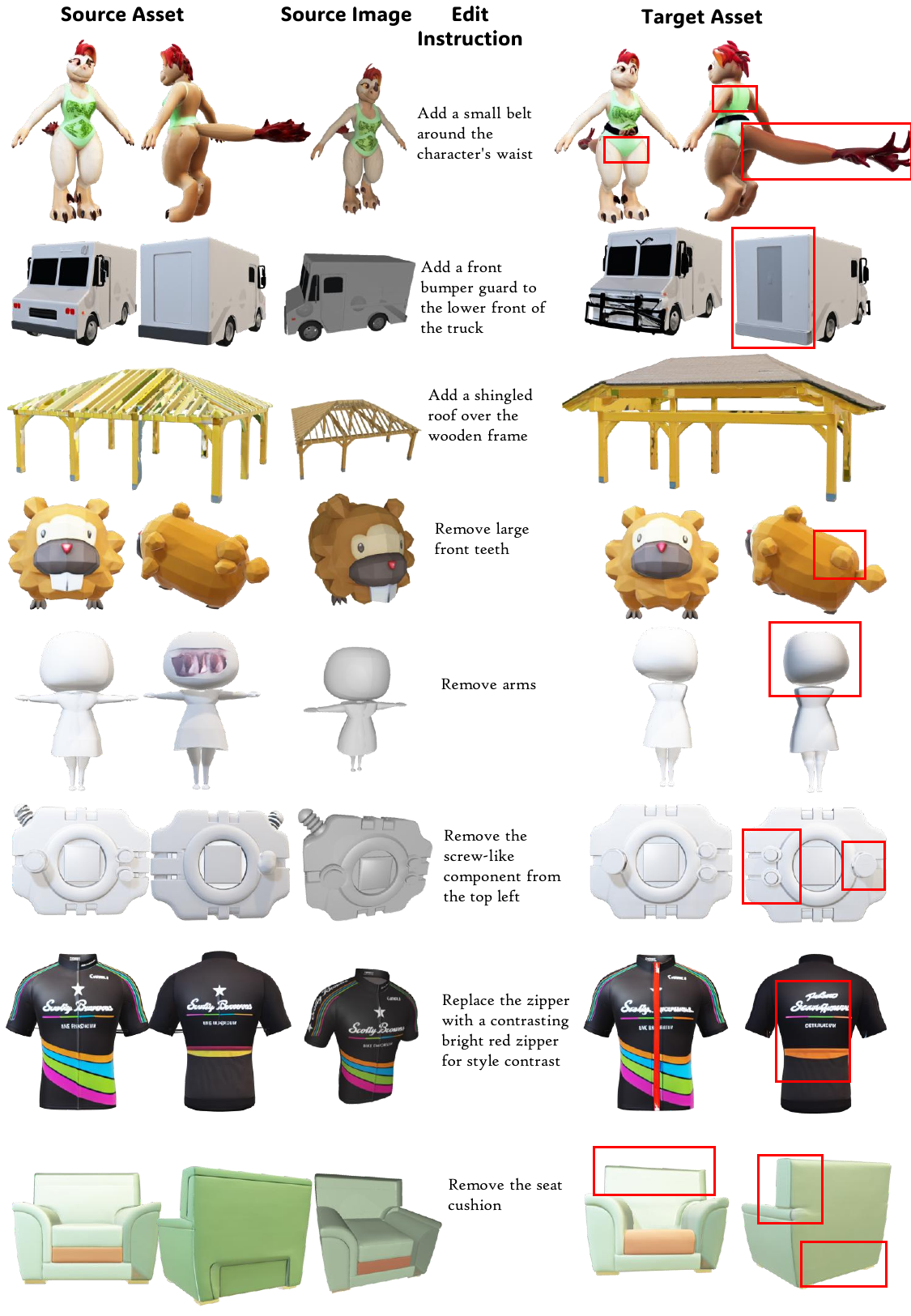}
\end{center}
\caption{Examples from the Edit3D-Bench of Steer3D.}
\label{fig:steer3d_bench_vis}
\end{figure}

\begin{figure}[h]
\begin{center}
\includegraphics[width=\linewidth]{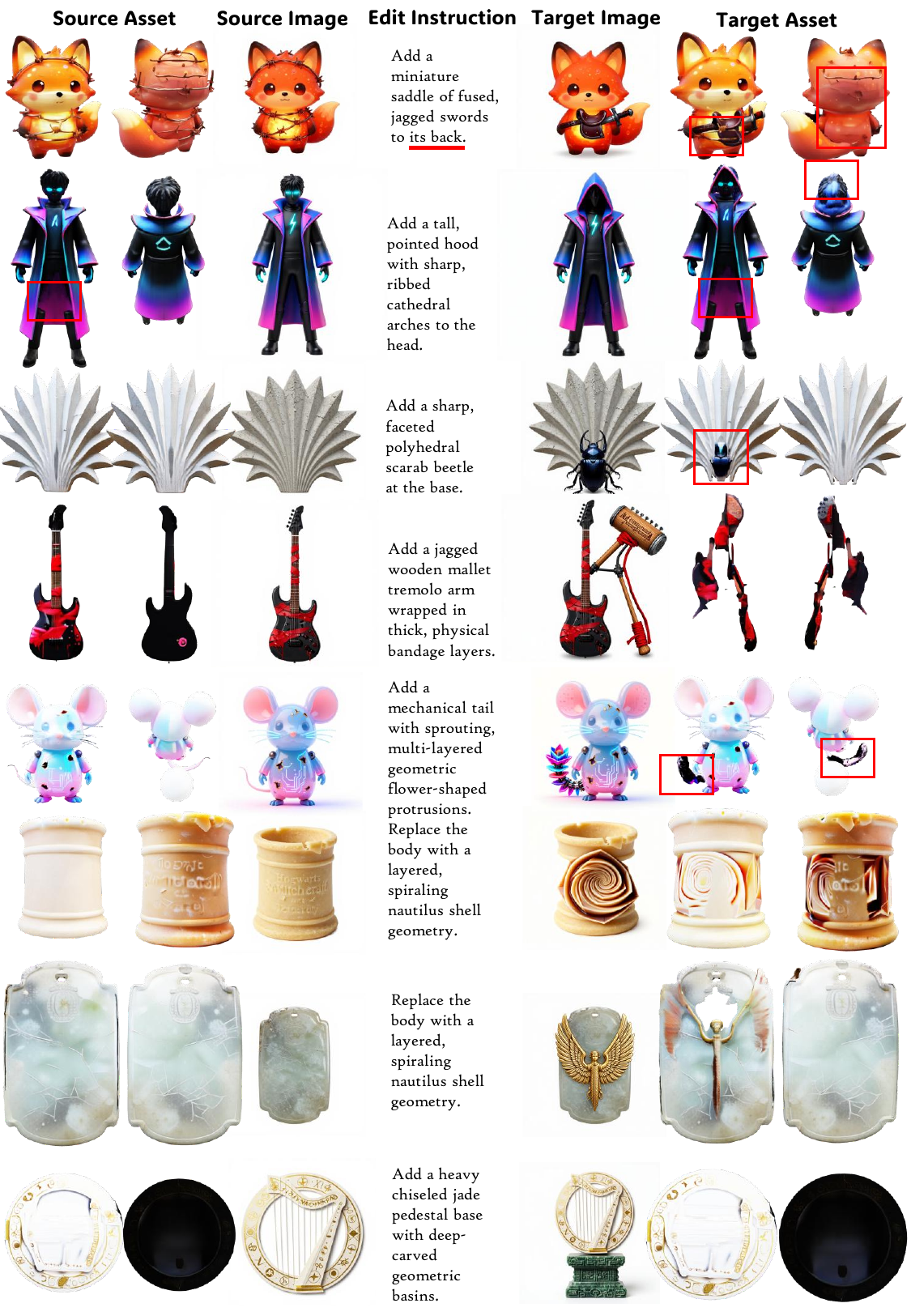}
\end{center}
\caption{Examples from Nano3D-100K. Nano3D did not release its test set, so this figure uses the 1.5K-pair subset described in Sec.~\ref{subsubsec:app_image_results}.}
\label{fig:nano3d_bench_vis}
\end{figure}

\begin{figure}[h]
\begin{center}
\includegraphics[width=\linewidth]{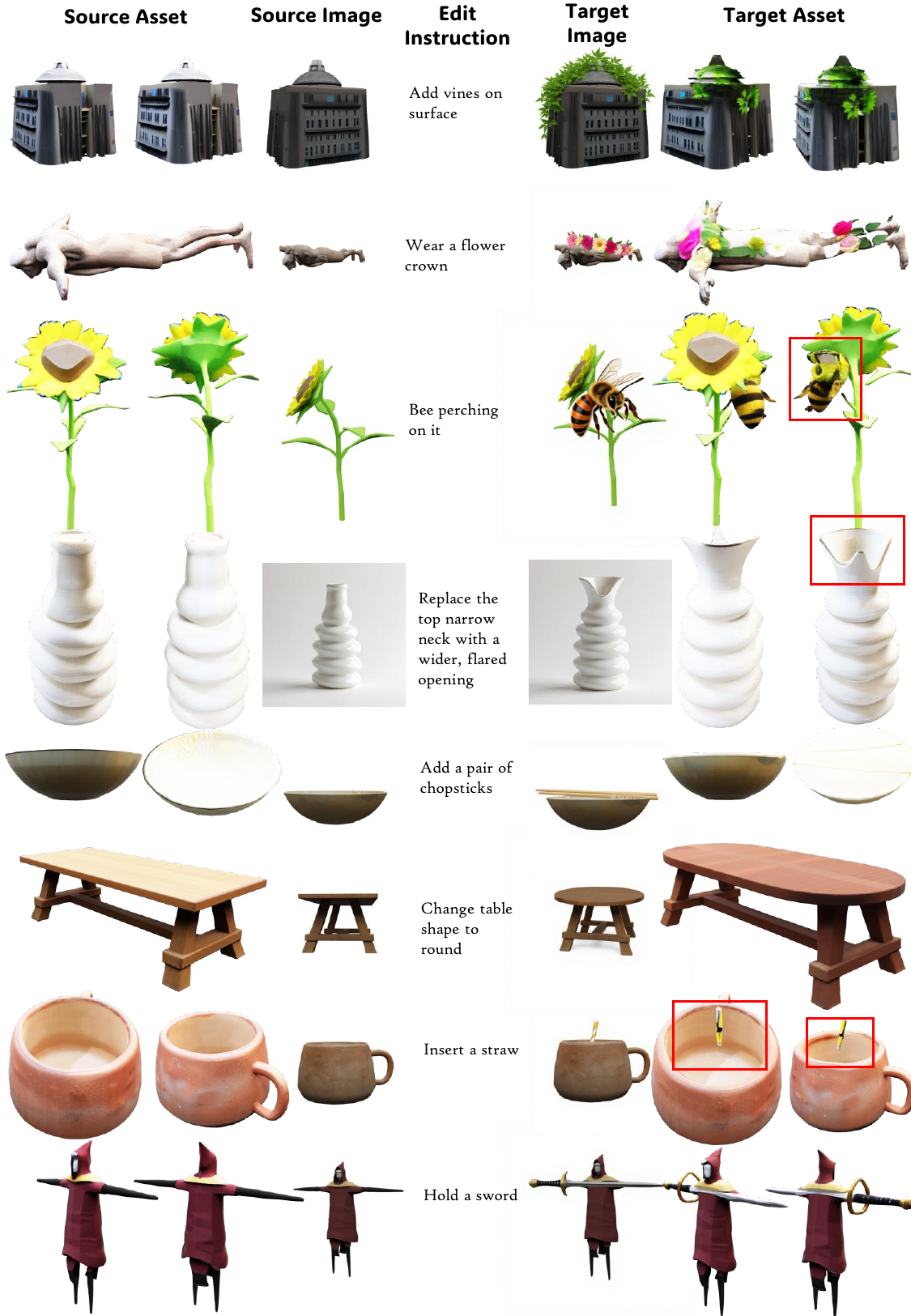}
\end{center}
\caption{Examples from the 3DEditVerse test set.}
\label{fig:3deditverse_bench_vis}
\end{figure}

We provide visualizations of samples for the Edit3D-Bench proposed by Steer3D in Fig.~\ref{fig:steer3d_bench_vis}, for Nano3D-100K in Fig.~\ref{fig:nano3d_bench_vis}, and for 3DEditVerse in Fig.~\ref{fig:3deditverse_bench_vis}.

\paragraph{Insufficient Evaluation Dimensions.}
Existing evaluation protocols also provide limited coverage of the diverse requirements of 3D asset editing. In particular, measuring similarity to an edited reference alone does not fully characterize whether a model successfully executes the requested modification, preserves the identity and irrelevant attributes of the source asset, or produces a perceptually high-quality result. A comprehensive benchmark should therefore jointly assess multiple aspects of editing quality, including editing success, instruction following, source identity preservation, visual quality, and semantic alignment with the desired edit.

\subsubsection{Construction Pipeline}
\label{subsubsec:gedit3d_construction}
Each sample in GEdit3D-Bench consists of a source 3D asset, a rendered source image, a natural-language editing instruction, and a corresponding target edited image, along with the text captions for source and target assets. Given a source asset $\mathcal{A}$ and an editing instruction $\mathcal{I}$ or edited image $\mathcal{I}_{\text{tgt}}$, a 3D editing model is expected to generate an edited asset $\mathcal{A}'$ that satisfies the requested modification while preserving the identity and irrelevant attributes of the source asset. The construction pipeline consists of four stages.

\paragraph{3D Asset Synthesis and Collection.}
We first define 21 high-level asset categories covering a broad range of semantic domains, including characters, animals, vehicles, furniture, architecture, and daily objects. For each category, Gemini-3.5-Flash\footnote{\url{https://deepmind.google/models/model-cards/gemini-3-5-flash/}} generates diverse asset descriptions, which are subsequently converted into high-quality images using Cosmos3-Super-Text2Image~\citep{cosmos3}. These images are then transformed into 3D assets using Hunyuan3D V3.1. To further improve the diversity and realism of the benchmark, we additionally collect real-world 3D assets from Sketchfab. During collection, we consider semantic categories, user engagement, and release timestamps. By prioritizing recently released assets, we reduce potential overlap with existing large-scale 3D generation datasets, such as Objaverse-XL~\citep{objaversexl} and TexVerse~\citep{texverse}. Combining generated and real-world assets enables GEdit3D-Bench to cover a broad spectrum of use cases, object categories, visual styles, and geometric structures.

\paragraph{3D Asset Captioning.}
For each collected asset, we render multiple predefined viewpoints using Blender and concatenate them into multi-view image grids. These rendered views are provided to Gemini-3.5-Flash to generate detailed semantic descriptions of the assets. The resulting captions describe object categories, appearances, materials, structures, and distinctive characteristics, and serve as semantic references for subsequent instruction generation.

\paragraph{Editing Instruction and Target Image Generation.}
Given each source asset, we randomly sample rendered viewpoints and provide them together with the source captions to Gemini-3.5-Flash. Using category-specific prompting templates, Gemini-3.5-Flash generates diverse editing instructions covering addition, removal, replacement, local and global appearance modification, and animation. For each instruction, HunyuanImage3.0-Instruct~\citep{hunyuanimage3} performs image-level editing on the rendered source views to generate corresponding target images. These edited images provide visual references for evaluating whether a 3D editing model correctly interprets and executes the requested modification.

\paragraph{Multi-stage Quality Curation.}
To ensure the reliability of benchmark samples, each editing triplet, consisting of a source rendering, editing instruction, and edited image, undergoes multi-stage quality verification. Specifically, we employ Gemini-3.5-Flash, GPT-5.6-Sol, and human experts to filter samples with incorrect semantics, unrealistic modifications, inconsistent object identities, or low-quality editing results. After quality control, we randomly sample the remaining high-quality triplets to form the final benchmark set. Finally, source captions and editing instructions are provided to GPT-5.6-Sol to generate target captions describing the expected edited assets.

\subsubsection{Benchmark Statistics}
\begin{figure}[h]
\begin{center}
\includegraphics[width=0.95\linewidth]{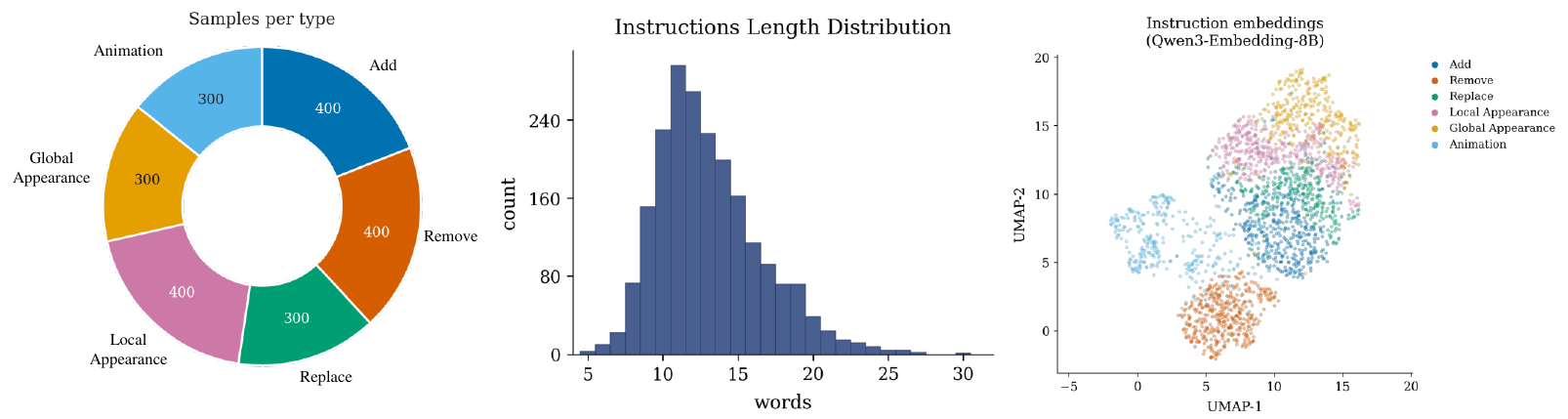}
\end{center}
\caption{Statistics of GEdit3D-Bench. The left panel shows the number of samples per editing type, the middle panel shows the distribution of instruction length, and the right panel visualizes instruction embeddings from Qwen3-Embedding-8B.}
\label{fig:bench_stats}
\end{figure}

As shown in Fig.~\ref{fig:bench_stats}, GEdit3D-Bench contains 400 samples for addition, removal, and local appearance, and 300 samples for replacement, animation, and global appearance. The editing instructions are about a dozen words long, as in our training set, and describe short atomic edits. We embed these instructions with Qwen3-Embedding-8B\footnote{\url{https://huggingface.co/Qwen/Qwen3-Embedding-8B}} and visualize them with UMAP in the right panel of Fig.~\ref{fig:bench_stats}. The embedding distribution indicates that the instructions cover diverse edits. 

\subsubsection{Evaluation Protocol Details}
\label{subsubsec:gedit3d_protocol}
\paragraph{View Quality Assessment.}
To assess the perceptual quality of edited assets, we render multiple views and compute image-quality scores for each view. Specifically, we employ the aesthetic-predictor-v2.5\footnote{\url{https://github.com/discus0434/aesthetic-predictor-v2-5}} and established image quality assessment models, including MANIQA~\citep{maniqa} and MUSIQ~\citep{musiq}, implemented through pyiqa~\citep{pyiqa}. Scores are averaged across the rendered views to obtain an overall view-quality assessment.

\paragraph{Reference Alignment.}
Pretrained encoders are widely used to evaluate semantic alignment between generated results and desired targets. We consider image alignment $A_I$ and text alignment $A_T$, instantiated with EVA-CLIP~\citep{evaclip}, SigLIP~\citep{siglip}, BLIP~\citep{blip}, DINOv3~\citep{dinov3}, and Uni3D~\citep{uni3d}, where applicable.


\paragraph{MLLM-based Evaluation.}
Recent benchmarks for video and image generation and editing~\citep{point2insert,goku,imgedit,vbench} have increasingly adopted modern multimodal large language models (MLLMs) for evaluation due to their strong visual understanding and reasoning capabilities. Following this direction, we instruct MLLMs to evaluate each editing result along four complementary dimensions: \textbf{Success Rate (SR)}, \textbf{Instruction Following (IF)}, \textbf{Identity Preservation (IP)}, and \textbf{Visual Quality (VQ)}. SR is a binary judgment of whether the requested edit is successfully achieved, while IF, IP, and VQ are scored on a 1--100 scale. Specifically, IF measures how faithfully the result follows the editing instruction, IP measures whether the identity and irrelevant properties of the source asset are preserved, and VQ measures the perceptual quality and plausibility of the resulting asset. SR and VQ use a shared system prompt. IF and IP use an edit-type-specific prompt, because the editing types differ substantially in granularity.

\subsection{More Evaluation Details and Results}
\label{subsec:app_eval}
\subsubsection{Details of Evaluation Setting}
\label{subsubsec:exp_details}
\paragraph{Main experiments.}
Unless otherwise stated, the CLIP model used in evaluation is EVA-CLIP-18B\footnote{\url{https://huggingface.co/BAAI/EVA-CLIP-18B}}, the SigLIP model is siglip2-giant\footnote{\url{https://huggingface.co/google/siglip2-giant-opt-patch16-384}}, the DINO model is DINOv3 ViT-L\footnote{\url{https://huggingface.co/facebook/dinov3-vitl16-pretrain-lvd1689m}}, and the Uni3D model is uni3d-giant\footnote{\url{https://huggingface.co/BAAI/Uni3D/tree/main/modelzoo/uni3d-g}}.
For reference alignment, each asset is rendered from ten views with camera radius 1.8 and a $49^{\circ}$ field of view. Eight views use a pitch of $10^{\circ}$ and yaw angles spaced by $45^{\circ}$. The other two are front views, with yaw fixed to the front and pitch set to $+30^{\circ}$ and $-30^{\circ}$. MLLM scores use the same random 400-sample subset of GEdit3D-Bench as the user study.

\paragraph{Instruction-driven editing by Alchemy3D.}
Noisy instruction captions make Alchemy3D-Instruct weaker than image-conditioned Alchemy3D, a gap also observed in text-conditioned 3D generation~\citep{trellis}. We therefore design a complementary pipeline that uses image-conditioned Alchemy3D for instruction-driven editing. Given a source asset and an editing instruction, we render 16 random views and ask a VLM to select the view best suited to the requested edit. A 2D image editor modifies that view, after which Alchemy3D edits the source asset using the modified image as its reference. We use Qwen3.8-27B\footnote{\url{https://huggingface.co/Qwen/Qwen3.8-27B}} for view selection and FLUX.2-Klein-9B\footnote{\url{https://huggingface.co/black-forest-labs/FLUX.2-klein-9B}} for image editing.

\begin{figure}[h]
\begin{center}
\includegraphics[width=0.8\linewidth]{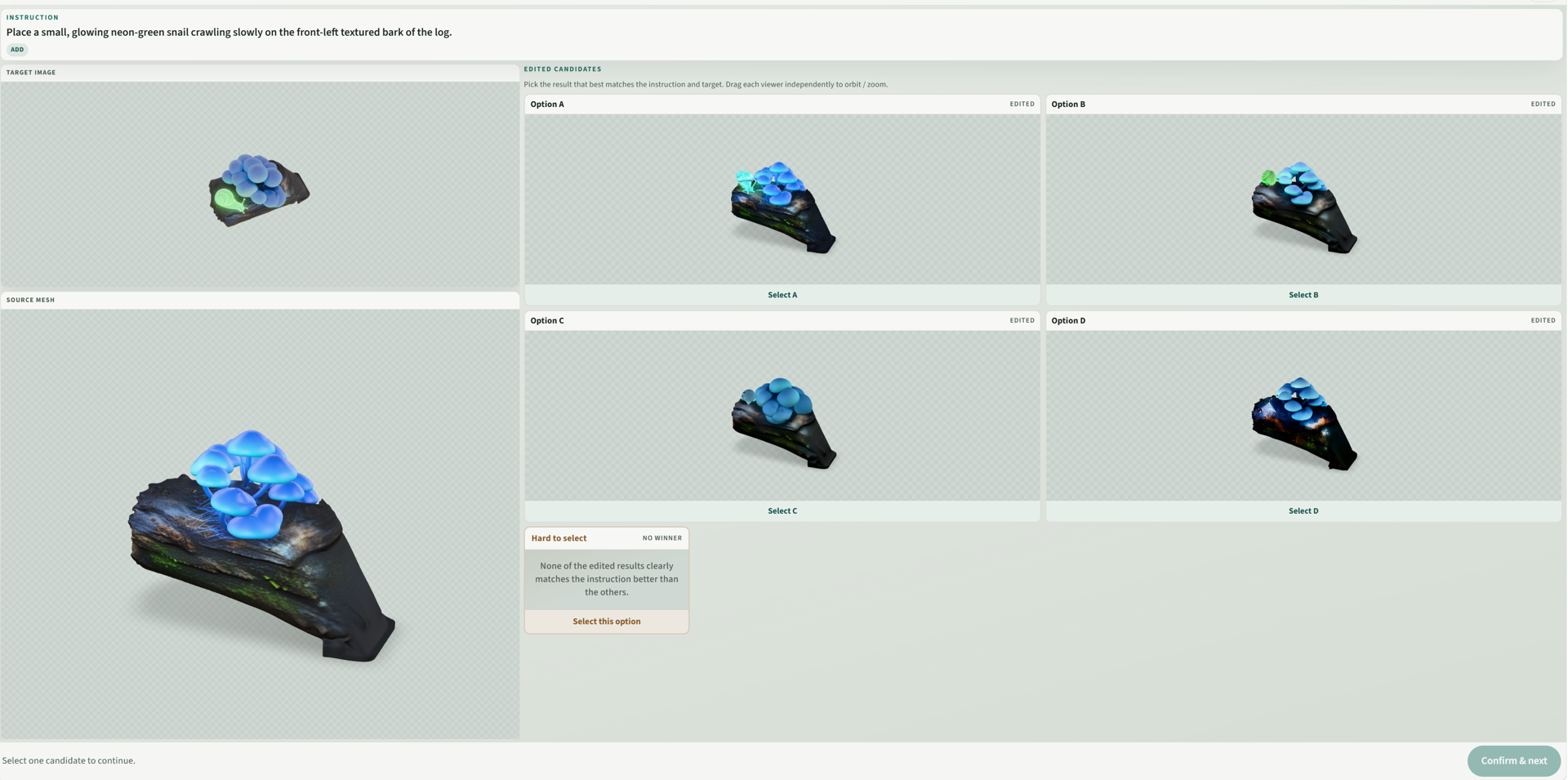}
\end{center}
\caption{Example page from the user study.}
\label{fig:user_study}
\end{figure}

\paragraph{User Study.}
To evaluate the methods in a realistic setting, we conducted a user study with 24 participants from nine institutions. We used the same subset as for MLLM scoring and randomly assigned 50 tasks to each participant. For each task, the source asset and outputs from all methods were loaded into Google Model Viewer\footnote{\url{https://github.com/google/model-viewer}}, allowing participants to inspect each asset interactively. Participants selected the best result or chose ``Hard to Select'', which we counted as ``Cannot Decide''. An example page is shown in Fig.~\ref{fig:user_study}.

\subsection{More Experimental Results}

\begin{figure}[h]
\begin{center}
\includegraphics[width=0.95\linewidth]{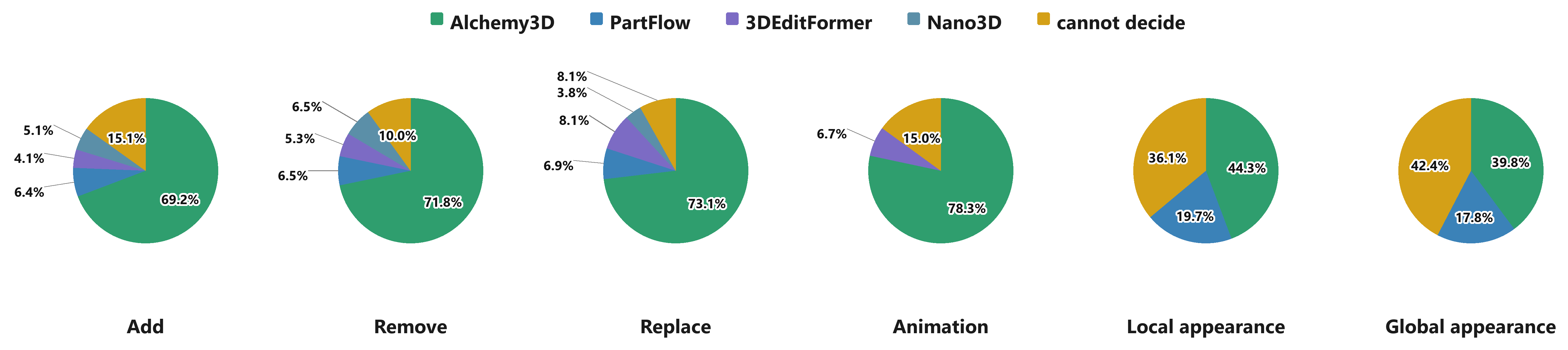}
\end{center}
\caption{Results of the user study.}
\label{fig:user_result}
\end{figure}

Fig.~\ref{fig:user_result} reports user-selection rates across all six editing types. Alchemy3D receives the highest preference for all editing types: addition (69.2\%), removal (71.8\%), replacement (73.1\%), animation (78.3\%), local appearance (44.3\%), and global appearance (39.8\%). The ``Cannot Decide'' rate increases from 8.1--15.1\% for structural and animation edits to 36.1\% and 42.4\% for local and global appearance edits, respectively, indicating greater ambiguity in evaluating appearance changes. 

\subsubsection{Image-conditioned Editing}
\label{subsubsec:app_image_results}
We first report complete image-conditioned results on Eval3DEdit and GEdit3D-Bench, followed by evaluations on 3DEditVerse, the Edit3D-Bench of VoxHammer, and Nano3D-100K. Nano3D supports only addition, removal, and replacement. 3DEditFormer does not support appearance edits, and PartFlow does not support animation edits. We evaluate each baseline only on the editing types it supports. 

\begin{table*}[t]
    \centering
    \caption{Quantitative comparison on GEdit3D-Bench. Best results are in \textbf{bold} and second-best are \underline{underlined}.}
    \label{tab:gedit3d_full}
    \scriptsize
    \setlength{\tabcolsep}{2.0pt}
    \renewcommand{\arraystretch}{0.92}
    \resizebox{\linewidth}{!}{%
    \begin{tabular}{lccccccccccccc}
        \toprule
        & \multicolumn{3}{c}{\textbf{View Quality}}
        & \multicolumn{6}{c}{\textbf{Ref.\ Alignment}}
        & \multicolumn{4}{c}{\textbf{MLLM}} \\
        \cmidrule(lr){2-4}
        \cmidrule(lr){5-10}
        \cmidrule(lr){11-14}
        \textbf{Method}
        & Aes.\ $\uparrow$
        & MANIQA $\uparrow$
        & MUSIQ $\uparrow$
        & $A_I^{\mathrm{CLIP}}$ $\uparrow$
        & $A_T^{\mathrm{CLIP}}$ $\uparrow$
        & $A_I^{\mathrm{SigLIP}}$ $\uparrow$
        & $A_I^{\mathrm{DINO}}$ $\uparrow$
        & $A_I^{\mathrm{Uni3D}}$ $\uparrow$
        & $A_T^{\mathrm{Uni3D}}$ $\uparrow$
        & SR $\uparrow$
        & VQ $\uparrow$
        & IF $\uparrow$
        & IP $\uparrow$ \\
        \midrule

        \rowcolor{gray!12}
        \multicolumn{14}{l}{\textit{add}} \\
        Nano3D
        & 4.40 & \underline{0.552} & \underline{73.04} & 80.11 & 65.34 & 86.96 & 66.70 & \underline{36.48} & \underline{80.70} & 56.2\% & \underline{70.1} & 55.3 & \underline{72.5} \\
        3DEditFormer
        & 4.36 & 0.537 & 70.91 & 79.98 & 63.44 & 86.17 & 65.01 & 36.06 & 79.55 & 53.8\% & 63.8 & 56.1 & 59.4 \\
        PartFlow
        & 4.32 & 0.543 & 70.98 & 79.89 & 62.74 & 86.50 & 65.20 & 36.21 & 79.34 & 57.5\% & 63.4 & 56.5 & 65.8 \\
        \textbf{Alchemy3D-Turbo}
        & \textbf{4.59} & 0.541 & \textbf{73.49} & \underline{82.01} & \textbf{69.66} & \underline{88.59} & \underline{70.80} & 36.24 & 80.32 & \underline{85.0\%} & 68.6 & \underline{68.3} & 71.8 \\
        \textbf{Alchemy3D}
        & \underline{4.47} & \textbf{0.578} & 72.94 & \textbf{84.84} & \underline{68.66} & \textbf{90.28} & \textbf{71.63} & \textbf{36.87} & \textbf{81.31} & \textbf{86.2\%} & \textbf{72.2} & \textbf{69.8} & \textbf{76.3} \\

        \midrule
        \rowcolor{gray!12}
        \multicolumn{14}{l}{\textit{remove}} \\
        Nano3D
        & 4.39 & \underline{0.553} & \textbf{72.44} & 82.53 & 64.12 & 89.28 & 70.39 & \textbf{37.32} & \underline{81.05} & 41.2\% & 70.3 & 48.8 & 70.5 \\
        3DEditFormer
        & 4.33 & 0.548 & 70.69 & \underline{83.27} & 61.80 & 89.20 & 69.93 & 37.06 & 79.63 & \underline{52.5\%} & 68.3 & 58.0 & 65.9 \\
        PartFlow
        & 4.30 & 0.550 & 70.73 & 82.58 & 61.05 & 89.08 & 69.25 & 36.97 & 79.44 & 50.0\% & 66.1 & 53.5 & 65.1 \\
        \textbf{Alchemy3D-Turbo}
        & \textbf{4.56} & 0.535 & \underline{72.26} & 82.19 & \textbf{66.77} & \underline{89.54} & \textbf{74.28} & 36.66 & 80.61 & \textbf{90.0\%} & \underline{70.7} & \underline{71.7} & \underline{73.9} \\
        \textbf{Alchemy3D}
        & \underline{4.41} & \textbf{0.572} & 71.64 & \textbf{84.85} & \underline{65.93} & \textbf{91.23} & \underline{73.67} & \underline{37.15} & \textbf{81.24} & \textbf{90.0\%} & \textbf{73.0} & \textbf{72.8} & \textbf{75.6} \\

        \midrule
        \rowcolor{gray!12}
        \multicolumn{14}{l}{\textit{replace}} \\
        Nano3D
        & 4.41 & \underline{0.557} & \textbf{73.20} & 82.29 & 66.16 & 88.28 & 70.75 & \textbf{37.69} & \underline{82.91} & 57.6\% & 70.0 & 58.7 & 73.7 \\
        3DEditFormer
        & 4.34 & 0.545 & 71.06 & 82.38 & 64.77 & 88.19 & 69.66 & \underline{37.68} & 82.83 & 74.6\% & 67.0 & 67.2 & 65.4 \\
        PartFlow
        & 4.32 & 0.548 & 71.23 & 81.29 & 63.28 & 88.14 & 69.28 & 37.32 & 81.32 & 60.3\% & 66.7 & 59.6 & 69.7 \\
        \textbf{Alchemy3D-Turbo}
        & \textbf{4.58} & 0.540 & \underline{73.11} & \underline{83.05} & \textbf{68.91} & \underline{89.62} & \underline{73.80} & 37.13 & 82.36 & \underline{90.0\%} & \underline{70.6} & \textbf{72.6} & \underline{74.2} \\
        \textbf{Alchemy3D}
        & \underline{4.43} & \textbf{0.578} & 72.53 & \textbf{85.10} & \underline{68.22} & \textbf{90.67} & \textbf{74.60} & 37.41 & \textbf{82.99} & \textbf{93.3\%} & \textbf{73.2} & \underline{71.5} & \textbf{76.8} \\

        \midrule
        \rowcolor{gray!12}
        \multicolumn{14}{l}{\textit{local\_appearance}} \\
        PartFlow
        & 4.32 & \underline{0.552} & 70.53 & \underline{81.67} & 61.36 & 89.21 & 71.31 & \textbf{36.76} & 79.42 & 38.8\% & 69.7 & 45.2 & 68.5 \\
        \textbf{Alchemy3D-Turbo}
        & \textbf{4.60} & 0.541 & \textbf{72.93} & 81.30 & \textbf{66.73} & \underline{89.30} & \textbf{73.60} & 36.45 & \underline{80.43} & \textbf{61.3\%} & \underline{70.0} & \textbf{57.4} & \underline{75.0} \\
        \textbf{Alchemy3D}
        & \underline{4.44} & \textbf{0.579} & \underline{71.97} & \textbf{83.45} & \underline{65.56} & \textbf{90.44} & \underline{72.78} & \underline{36.64} & \textbf{80.88} & \underline{54.4\%} & \textbf{72.4} & \underline{53.4} & \textbf{76.0} \\

        \midrule
        \rowcolor{gray!12}
        \multicolumn{14}{l}{\textit{global\_appearance}} \\
        PartFlow
        & 4.36 & \underline{0.542} & 70.14 & 72.04 & 55.31 & 82.01 & 64.38 & 34.20 & 73.50 & 8.3\% & \underline{68.7} & 36.1 & 75.0 \\
        \textbf{Alchemy3D-Turbo}
        & \textbf{4.56} & 0.502 & \textbf{71.20} & \underline{74.57} & \textbf{63.17} & 83.62 & \underline{67.09} & \textbf{34.51} & \textbf{75.35} & \textbf{50.0\%} & 68.3 & \textbf{53.2} & \underline{76.8} \\
        \textbf{Alchemy3D}
        & \underline{4.45} & \textbf{0.565} & \underline{71.05} & \textbf{75.72} & 60.79 & \textbf{84.46} & \textbf{67.82} & \underline{34.36} & \underline{74.99} & \underline{28.3\%} & \textbf{72.7} & \underline{43.7} & \textbf{77.7} \\

        \midrule
        \rowcolor{gray!12}
        \multicolumn{14}{l}{\textit{animation}} \\
        3DEditFormer
        & 4.19 & \underline{0.535} & 70.59 & \underline{83.83} & 65.88 & 88.91 & 70.60 & \underline{36.79} & 79.38 & 49.2\% & 63.9 & 54.9 & 61.6 \\
        \textbf{Alchemy3D-Turbo}
        & \textbf{4.42} & 0.533 & \textbf{72.78} & 83.62 & \textbf{69.97} & \underline{89.96} & \underline{73.88} & 36.44 & \underline{80.59} & \textbf{68.3\%} & \underline{64.8} & \underline{65.1} & \underline{69.2} \\
        \textbf{Alchemy3D}
        & \underline{4.31} & \textbf{0.567} & \underline{72.34} & \textbf{86.41} & \underline{69.29} & \textbf{91.44} & \textbf{75.23} & \textbf{37.31} & \textbf{82.02} & \underline{66.7\%} & \textbf{70.8} & \textbf{66.6} & \textbf{72.8} \\

        \bottomrule
    \end{tabular}%
    }
\end{table*}

On addition, removal, and replacement in Table~\ref{tab:gedit3d_full}, Alchemy3D and Alchemy3D-Turbo lead in success rate, instruction following, and most alignment scores. Removal is the clearest case: both of our models reach 90\% success, against 41.2\% for Nano3D and about 50\% for 3DEditFormer and PartFlow. Animation follows the same ranking. Alchemy3D-Turbo reaches 68.3\% success and Alchemy3D 66.7\%, against 49.2\% for 3DEditFormer, and Alchemy3D is also highest on image alignment and identity preservation.

Appearance edits remain weaker than structural edits. PartFlow reaches 38.8\% success on local appearance and 8.3\% on global appearance. Alchemy3D-Turbo reaches 61.3\% and 50.0\%, and Alchemy3D reaches 54.4\% and 28.3\%, while both preserve identity better than PartFlow. We attribute the remaining gap to task conflict under joint training: most training pairs modify both geometry and appearance, so appearance-only edits conflict with the majority of the training distribution (Sec.~\ref{subsec:app_future}).

\begin{table*}[t]
    \centering
    \caption{Quantitative comparison on Eval3DEdit. Best results are in \textbf{bold} and second-best are \underline{underlined}.}
    \label{tab:eval3dedit_full}
    \scriptsize
    \setlength{\tabcolsep}{2.0pt}
    \renewcommand{\arraystretch}{0.92}
    \resizebox{\linewidth}{!}{%
    \begin{tabular}{lccccccccccccc}
        \toprule
        & \multicolumn{3}{c}{\textbf{View Quality}}
        & \multicolumn{6}{c}{\textbf{Ref.\ Alignment}}
        & \multicolumn{4}{c}{\textbf{MLLM Score}} \\
        \cmidrule(lr){2-4}
        \cmidrule(lr){5-10}
        \cmidrule(lr){11-14}
        \textbf{Method}
        & Aes.\ $\uparrow$
        & MANIQA $\uparrow$
        & MUSIQ $\uparrow$
        & $A_I^{\mathrm{CLIP}}$ $\uparrow$
        & $A_T^{\mathrm{CLIP}}$ $\uparrow$
        & $A_I^{\mathrm{SigLIP}}$ $\uparrow$
        & $A_I^{\mathrm{DINO}}$ $\uparrow$
        & $A_I^{\mathrm{Uni3D}}$ $\uparrow$
        & $A_T^{\mathrm{Uni3D}}$ $\uparrow$
        & SR $\uparrow$
        & VQ $\uparrow$
        & IF $\uparrow$
        & IP $\uparrow$ \\
        \midrule

        \rowcolor{gray!12}
        \multicolumn{14}{l}{\textit{add}} \\
        Nano3D
        & 4.65 & 0.547 & 72.09 & 80.58 & 46.70 & 87.89 & 70.46 & 37.94 & 61.59 & 27.3\% & \underline{68.9} & 38.3 & 66.9 \\
        3DEditFormer
        & 4.57 & 0.523 & 72.05 & 79.49 & 45.56 & 86.72 & 67.06 & 37.29 & 59.88 & 45.5\% & 66.2 & 45.8 & 62.3 \\
        PartFlow
        & 4.60 & \underline{0.549} & 72.25 & 80.63 & 48.47 & 87.37 & 68.44 & 36.72 & 60.96 & 62.5\% & 63.0 & 55.0 & 69.9 \\
        \textbf{Alchemy3D-Turbo}
        & \textbf{4.75} & 0.537 & \textbf{74.80} & \underline{81.99} & \underline{53.05} & \underline{88.93} & \underline{74.61} & \underline{38.80} & \underline{66.20} & \underline{80.0\%} & 68.4 & \textbf{64.5} & \textbf{77.0} \\
        \textbf{Alchemy3D}
        & \underline{4.68} & \textbf{0.559} & \underline{74.08} & \textbf{83.21} & \textbf{53.06} & \textbf{89.70} & \textbf{74.89} & \textbf{38.86} & \textbf{66.28} & \textbf{81.8\%} & \textbf{74.1} & \underline{63.7} & \underline{72.5} \\

        \midrule
        \rowcolor{gray!12}
        \multicolumn{14}{l}{\textit{remove}} \\
        Nano3D
        & \textbf{4.65} & 0.543 & 72.37 & \textbf{84.27} & \textbf{49.12} & \underline{90.48} & 75.47 & 40.90 & \textbf{62.51} & 20.0\% & 66.2 & 37.8 & \underline{74.6} \\
        3DEditFormer
        & 4.57 & 0.518 & 71.46 & 82.94 & \underline{47.98} & 89.35 & 73.31 & 39.99 & 58.14 & 70.0\% & 65.0 & 60.7 & 67.8 \\
        PartFlow
        & 4.49 & 0.531 & 71.45 & 82.45 & 47.10 & 89.05 & 70.25 & 39.91 & 58.71 & \underline{80.0\%} & 62.9 & 66.2 & 69.9 \\
        \textbf{Alchemy3D-Turbo}
        & \underline{4.64} & \underline{0.554} & \textbf{74.40} & 83.09 & 47.09 & 89.93 & \underline{76.70} & \textbf{41.06} & \underline{60.66} & \textbf{90.0\%} & \textbf{69.5} & \underline{70.6} & \textbf{77.1} \\
        \textbf{Alchemy3D}
        & 4.59 & \textbf{0.560} & \underline{72.73} & \underline{84.05} & 46.69 & \textbf{90.94} & \textbf{78.12} & \underline{40.95} & 59.61 & \underline{80.0\%} & \underline{69.3} & \textbf{71.0} & 72.0 \\

        \midrule
        \rowcolor{gray!12}
        \multicolumn{14}{l}{\textit{replace}} \\
        Nano3D
        & 4.51 & 0.542 & 72.20 & 82.06 & 52.64 & 88.20 & 73.65 & 38.73 & 67.49 & 77.8\% & \underline{69.0} & 64.3 & \underline{74.8} \\
        3DEditFormer
        & 4.50 & 0.535 & 71.93 & 82.90 & 54.18 & 88.49 & 72.89 & 38.66 & 66.76 & \textbf{100.0\%} & 65.1 & \textbf{74.0} & 73.6 \\
        PartFlow
        & 4.43 & \underline{0.542} & 72.44 & 81.70 & 52.37 & 88.28 & 72.77 & 38.77 & 68.68 & 60.0\% & 67.1 & 58.7 & 71.9 \\
        \textbf{Alchemy3D-Turbo}
        & \underline{4.54} & 0.536 & \textbf{73.94} & \underline{82.95} & \textbf{57.54} & \underline{88.84} & \underline{74.72} & \underline{39.62} & \textbf{71.84} & \textbf{100.0\%} & \textbf{71.2} & \underline{73.2} & \textbf{76.4} \\
        \textbf{Alchemy3D}
        & \textbf{4.54} & \textbf{0.553} & \underline{73.46} & \textbf{84.04} & \underline{56.56} & \textbf{89.75} & \textbf{77.63} & \textbf{39.96} & \underline{71.72} & \underline{90.0\%} & 67.0 & 69.5 & 73.5 \\

        \midrule
        \rowcolor{gray!12}
        \multicolumn{14}{l}{\textit{style}} \\
        3DEditFormer
        & 4.61 & 0.507 & 70.84 & 81.05 & 51.72 & \underline{88.57} & 67.24 & \textbf{39.17} & \underline{66.35} & \underline{90.0\%} & \underline{67.2} & -- & -- \\
        PartFlow
        & 4.58 & \underline{0.532} & \underline{71.63} & 78.10 & 50.68 & 86.29 & 64.58 & 37.17 & 65.70 & 40.0\% & 61.7 & -- & -- \\
        \textbf{Alchemy3D-Turbo}
        & \underline{4.61} & 0.502 & 71.32 & \underline{81.15} & \underline{52.51} & 87.74 & \underline{71.97} & \underline{38.23} & 65.78 & \textbf{100.0\%} & 65.1 & -- & -- \\
        \textbf{Alchemy3D}
        & \textbf{4.64} & \textbf{0.552} & \textbf{72.53} & \textbf{82.40} & \textbf{54.41} & \textbf{89.89} & \textbf{73.34} & 38.16 & \textbf{67.47} & \underline{90.0\%} & \textbf{71.2} & -- & -- \\

        \midrule
        \rowcolor{gray!12}
        \multicolumn{14}{l}{\textit{action}} \\
        3DEditFormer
        & 4.06 & 0.509 & 70.32 & 75.80 & 44.45 & 83.56 & 65.69 & 32.86 & 54.34 & 20.0\% & 55.2 & 40.0 & 57.5 \\
        \textbf{Alchemy3D-Turbo}
        & \textbf{4.33} & \underline{0.513} & \textbf{73.58} & \underline{79.96} & \textbf{51.46} & \underline{86.85} & \underline{73.76} & \underline{36.85} & \underline{63.89} & \textbf{90.0\%} & \underline{64.4} & \textbf{70.5} & \underline{69.5} \\
        \textbf{Alchemy3D}
        & \underline{4.29} & \textbf{0.537} & \underline{72.83} & \textbf{80.84} & \underline{51.15} & \textbf{88.09} & \textbf{76.49} & \textbf{37.01} & \textbf{66.34} & \underline{80.0\%} & \textbf{70.8} & \textbf{70.5} & \textbf{75.6} \\

        \bottomrule
    \end{tabular}%
    }
\end{table*}

Table~\ref{tab:eval3dedit_full} separates Eval3DEdit by editing type. On addition, Alchemy3D reaches 81.8\% success, against 27.3\% for Nano3D, 45.5\% for 3DEditFormer, and 62.5\% for PartFlow. On removal, Nano3D keeps a high CLIP score but only 20\% success, so similarity to the source is not the same as completing the edit. Replacement is the closest comparison: 3DEditFormer and Alchemy3D-Turbo both reach 100\% success, while Alchemy3D is slightly lower at 90\% and remains stronger on image alignment. On style, PartFlow falls to 40\% success, whereas Alchemy3D-Turbo reaches 100\%. On action, 3DEditFormer succeeds on 20\% of cases, against 90\% for Alchemy3D-Turbo and 80\% for Alchemy3D.

We next evaluate on 3DEditVerse~\citep{3deditformer}, the Edit3D-Bench of VoxHammer~\citep{voxhammer}, and Nano3D-100K~\citep{nano3d}. Nano3D does not release an official test list, so we randomly sample 1.5K pairs. View quality and reference alignment follow Sec.~\ref{sec:benchmark} and do not use a reconstructed 3D target. For completeness, we also report \emph{target alignment}: perceptual similarity to rendered views of the provided target asset through LPIPS~\citep{lpips}, SSIM~\citep{ssim}, and PSNR, and geometric similarity to its mesh through Chamfer distance~\citep{cd}, F-score~\citep{f1}, and normal consistency~\citep{nc}. These columns are shaded. As discussed in Sec.~\ref{sec:benchmark}, a reconstructed asset is not an independent reference, and similarity to one such asset ignores the one-to-many nature of editing. Alchemy3D performs better on the unshaded metrics but worse on the shaded metrics, consistent with prior methods being trained or selected against the same reconstructed targets.

\begin{table*}[t]
    \centering
    \caption{Image-conditioned comparison on 3DEditVerse. Shaded columns measure alignment with the provided reconstructed target and are not treated as primary metrics.}
    \label{tab:3deditverse}
    \small
    \setlength{\tabcolsep}{3pt}
    \renewcommand{\arraystretch}{0.95}
    \resizebox{\linewidth}{!}{%
    \begin{tabular}{lccccccc>{\columncolor{gray!16}}c>{\columncolor{gray!16}}c>{\columncolor{gray!16}}c>{\columncolor{gray!16}}c>{\columncolor{gray!16}}c>{\columncolor{gray!16}}c}
        \toprule
        & \multicolumn{3}{c}{\textbf{View Quality}}
        & \multicolumn{4}{c}{\textbf{Ref.\ Alignment}}
        & \multicolumn{6}{c}{\cellcolor{gray!16}\textbf{Target Alignment}} \\
        \cmidrule(lr){2-4}
        \cmidrule(lr){5-8}
        \cmidrule(lr){9-14}
        \textbf{Method}
        & Aes.\ $\uparrow$ & MANIQA $\uparrow$ & MUSIQ $\uparrow$
        & $A_I^{\mathrm{CLIP}}$ $\uparrow$ & $A_I^{\mathrm{SigLIP}}$ $\uparrow$ & $A_I^{\mathrm{DINO}}$ $\uparrow$ & $A_I^{\mathrm{Uni3D}}$ $\uparrow$
        & LPIPS $\downarrow$ & SSIM $\uparrow$ & PSNR $\uparrow$
        & CD $\downarrow$ & F1 $\uparrow$ & NC $\uparrow$ \\
        \midrule
        3DEditFormer
        & 4.191 & 0.533 & 68.96
        & 69.18 & 79.18 & 64.60 & \underline{36.76}
        & \underline{0.0797} & \underline{0.9112} & \textbf{23.81}
        & \textbf{13.36} & \textbf{72.18} & \textbf{0.852} \\
        PartFlow
        & 4.145 & \underline{0.537} & 69.08
        & 68.08 & 78.60 & 63.74 & 36.46
        & \textbf{0.0786} & \textbf{0.9148} & \underline{23.75}
        & \underline{17.65} & \underline{66.92} & \underline{0.844} \\
        \textbf{Alchemy3D-Turbo}
        & \textbf{4.239} & 0.524 & \textbf{71.43}
        & \underline{72.10} & \underline{82.05} & \underline{69.44} & 36.36
        & 0.1016 & 0.8967 & 20.65
        & 22.71 & 50.28 & 0.773 \\
        \textbf{Alchemy3D}
        & \underline{4.213} & \textbf{0.558} & \underline{70.78}
        & \textbf{73.10} & \textbf{82.18} & \textbf{72.20} & \textbf{37.41}
        & 0.1000 & 0.8961 & 22.06
        & 24.48 & 44.15 & 0.757 \\
        \bottomrule
    \end{tabular}%
    }
\end{table*}

\begin{table}[t]
    \centering
    \caption{Image-conditioned comparison on Edit3D-Bench (VoxHammer). Target-alignment metrics are not available for this benchmark.}
    \label{tab:voxhammer}
    \small
    \setlength{\tabcolsep}{3pt}
    \renewcommand{\arraystretch}{0.95}
    \begin{tabular}{lccccccc}
        \toprule
        & \multicolumn{3}{c}{\textbf{View Quality}}
        & \multicolumn{4}{c}{\textbf{Ref.\ Alignment}} \\
        \cmidrule(lr){2-4}
        \cmidrule(lr){5-8}
        \textbf{Method}
        & Aes.\ $\uparrow$ & MANIQA $\uparrow$ & MUSIQ $\uparrow$
        & $A_I^{\mathrm{CLIP}}$ $\uparrow$ & $A_I^{\mathrm{SigLIP}}$ $\uparrow$ & $A_I^{\mathrm{DINO}}$ $\uparrow$ & $A_I^{\mathrm{Uni3D}}$ $\uparrow$ \\
        \midrule
        3DEditFormer
        & 4.099 & 0.553 & 73.72
        & 76.91 & 84.99 & 66.90 & 38.33 \\
        PartFlow
        & 4.112 & \underline{0.568} & \underline{74.25}
        & 77.55 & 85.95 & 69.02 & \underline{38.93} \\
        \textbf{Alchemy3D-Turbo}
        & \underline{4.124} & 0.553 & 73.48
        & \underline{79.90} & \underline{87.28} & \underline{72.91} & 38.25 \\
        \textbf{Alchemy3D}
        & \textbf{4.207} & \textbf{0.580} & \textbf{74.61}
        & \textbf{80.91} & \textbf{88.44} & \textbf{75.04} & \textbf{39.06} \\
        \bottomrule
    \end{tabular}
\end{table}

\begin{table*}[t]
    \centering
    \caption{Image-conditioned comparison on Nano3D-100K. Shaded columns measure alignment with the provided reconstructed target and are not treated as primary metrics.}
    \label{tab:nano3d}
    \small
    \setlength{\tabcolsep}{3pt}
    \renewcommand{\arraystretch}{0.95}
    \resizebox{\linewidth}{!}{%
    \begin{tabular}{lccccccc>{\columncolor{gray!16}}c>{\columncolor{gray!16}}c>{\columncolor{gray!16}}c>{\columncolor{gray!16}}c>{\columncolor{gray!16}}c>{\columncolor{gray!16}}c}
        \toprule
        & \multicolumn{3}{c}{\textbf{View Quality}}
        & \multicolumn{4}{c}{\textbf{Ref.\ Alignment}}
        & \multicolumn{6}{c}{\cellcolor{gray!16}\textbf{Target Alignment}} \\
        \cmidrule(lr){2-4}
        \cmidrule(lr){5-8}
        \cmidrule(lr){9-14}
        \textbf{Method}
        & Aes.\ $\uparrow$ & MANIQA $\uparrow$ & MUSIQ $\uparrow$
        & $A_I^{\mathrm{CLIP}}$ $\uparrow$ & $A_I^{\mathrm{SigLIP}}$ $\uparrow$ & $A_I^{\mathrm{DINO}}$ $\uparrow$ & $A_I^{\mathrm{Uni3D}}$ $\uparrow$
        & LPIPS $\downarrow$ & SSIM $\uparrow$ & PSNR $\uparrow$
        & CD $\downarrow$ & F1 $\uparrow$ & NC $\uparrow$ \\
        \midrule
        3DEditFormer
        & 4.228 & 0.500 & 71.27
        & 66.30 & 78.06 & 60.38 & 36.57
        & \underline{0.1084} & \underline{0.8740} & \underline{19.92}
        & \textbf{15.21} & \textbf{69.88} & \underline{0.814} \\
        PartFlow
        & 4.192 & \underline{0.509} & \underline{71.93}
        & 65.42 & 78.61 & 61.18 & 36.20
        & \textbf{0.0970} & \textbf{0.8887} & \textbf{20.79}
        & \underline{17.58} & \underline{69.43} & \textbf{0.823} \\
        \textbf{Alchemy3D-Turbo}
        & \underline{4.266} & 0.487 & 71.24
        & \underline{69.86} & \underline{80.81} & \underline{66.58} & \underline{37.04}
        & 0.1296 & 0.8603 & 18.79
        & 24.99 & 49.81 & 0.726 \\
        \textbf{Alchemy3D}
        & \textbf{4.304} & \textbf{0.539} & \textbf{73.27}
        & \textbf{72.22} & \textbf{82.37} & \textbf{71.11} & \textbf{38.06}
        & 0.1327 & 0.8572 & 18.40
        & 29.12 & 42.00 & 0.699 \\
        \bottomrule
    \end{tabular}%
    }
\end{table*}

Across Tables~\ref{tab:3deditverse}, \ref{tab:voxhammer}, and \ref{tab:nano3d}, Alchemy3D ranks first and Alchemy3D-Turbo usually second on the unshaded metrics.

\begin{table*}[t]
    \centering
    \caption{Instruction-driven comparison on GEdit3D-Bench. Steer3D does not support replacement or animation edits. $^{\dagger}$ Alchemy3D is evaluated with the agentic view selection and image editing described in Sec.~\ref{subsubsec:exp_details}.}
    \label{tab:gedit3d_instruct}
    \small
    \setlength{\tabcolsep}{3pt}
    \renewcommand{\arraystretch}{0.95}
    \begin{tabular}{lccccccccc}
        \toprule
        & \multicolumn{3}{c}{\textbf{View Quality}}
        & \multicolumn{2}{c}{\textbf{Ref.\ Alignment}}
        & \multicolumn{4}{c}{\textbf{MLLM}} \\
        \cmidrule(lr){2-4}
        \cmidrule(lr){5-6}
        \cmidrule(lr){7-10}
        \textbf{Method}
        & Aes.\ $\uparrow$ & MANIQA $\uparrow$ & MUSIQ $\uparrow$
        & $A_T^{\mathrm{CLIP}}$ $\uparrow$ & $A_I^{\mathrm{Uni3D}}$  $\uparrow$
        & SR $\uparrow$ & VQ $\uparrow$ & IF $\uparrow$ & IP $\uparrow$ \\
        \midrule
        \rowcolor{gray!12}
        \multicolumn{10}{l}{\textit{Add}} \\
        Steer3D
        & 3.810 & 0.522 & 69.39
        & 46.47 & 63.38
        & 0.0 & 47.9 & 25.7 & 30.2 \\
        Alchemy3D-Instruct
        & \underline{4.602} & \underline{0.563} & \underline{74.43}
        & \textbf{68.34} & \textbf{80.69}
        & \underline{57.5} & \underline{69.6} & \underline{57.9} & \textbf{75.5} \\
        Alchemy3D$^{\dagger}$
        & \textbf{4.604} & \textbf{0.566} & \textbf{74.48}
        & \underline{68.33} & \underline{78.76}
        & \textbf{75.0} & \textbf{72.4} & \textbf{65.6} & \underline{69.3} \\
        \midrule
        \rowcolor{gray!12}
        \multicolumn{10}{l}{\textit{Remove}} \\
        Steer3D
        & 3.827 & 0.523 & 69.18
        & 46.12 & 61.84
        & 8.8 & 53.0 & 41.7 & 32.9 \\
        Alchemy3D-Instruct
        & \underline{4.537} & \underline{0.549} & \underline{72.94}
        & \underline{65.15} & \underline{78.79}
        & \textbf{75.0} & \underline{66.8} & \underline{64.9} & \underline{68.3} \\
        Alchemy3D$^{\dagger}$
        & \textbf{4.590} & \textbf{0.557} & \textbf{73.11}
        & \textbf{66.34} & \textbf{80.69}
        & \underline{72.5} & \textbf{70.9} & \textbf{65.9} & \textbf{69.8} \\
        \midrule
        \rowcolor{gray!12}
        \multicolumn{10}{l}{\textit{Replace}} \\
        Alchemy3D-Instruct
        & \underline{4.443} & \underline{0.544} & \textbf{74.04}
        & \underline{65.99} & \underline{79.28}
        & \underline{78.1} & \underline{67.5} & \underline{67.1} & \underline{66.2} \\
        Alchemy3D$^{\dagger}$
        & \textbf{4.564} & \textbf{0.564} & \underline{73.98}
        & \textbf{67.78} & \textbf{81.85}
        & \textbf{87.1} & \textbf{70.8} & \textbf{71.2} & \textbf{68.8} \\
        \midrule
        \rowcolor{gray!12}
        \multicolumn{10}{l}{\textit{Local Appearance}} \\
        Steer3D
        & 3.840 & 0.528 & 69.54
        & 50.79 & 69.64
        & 23.8 & 55.1 & \underline{43.1} & 32.3 \\
        Alchemy3D-Instruct
        & \textbf{4.639} & \underline{0.558} & \underline{73.51}
        & \underline{65.10} & \underline{79.75}
        & \underline{30.0} & \underline{71.0} & 42.0 & \textbf{72.0} \\
        Alchemy3D$^{\dagger}$
        & \underline{4.604} & \textbf{0.563} & \textbf{73.52}
        & \textbf{65.72} & \textbf{80.03}
        & \textbf{52.5} & \textbf{71.2} & \textbf{53.2} & \underline{68.6} \\
        \midrule
        \rowcolor{gray!12}
        \multicolumn{10}{l}{\textit{Global Appearance}} \\
        Steer3D
        & 3.882 & 0.523 & 68.74
        & 49.79 & 68.78
        & \underline{18.3} & 53.3 & \underline{45.3} & 49.1 \\
        Alchemy3D-Instruct
        & \textbf{4.605} & \underline{0.538} & \textbf{72.88}
        & \underline{57.00} & \underline{71.86}
        & 6.7 & \underline{67.6} & 33.4 & \textbf{77.6} \\
        Alchemy3D$^{\dagger}$
        & \underline{4.580} & \textbf{0.540} & \underline{72.40}
        & \textbf{62.23} & \textbf{75.22}
        & \textbf{50.0} & \textbf{71.2} & \textbf{51.8} & \underline{75.5} \\
        \midrule
        \rowcolor{gray!12}
        \multicolumn{10}{l}{\textit{Animation}} \\
        Alchemy3D-Instruct
        & \underline{4.431} & \underline{0.549} & \underline{73.43}
        & \underline{69.20} & \underline{79.33}
        & \underline{46.7} & \underline{64.4} & \underline{56.2} & \underline{65.7} \\
        Alchemy3D$^{\dagger}$
        & \textbf{4.487} & \textbf{0.551} & \textbf{73.47}
        & \textbf{69.94} & \textbf{81.18}
        & \textbf{70.0} & \textbf{67.7} & \textbf{62.7} & \textbf{70.6} \\
        \bottomrule
    \end{tabular}
\end{table*}

\paragraph{Instruction-conditioned editing.}
\label{subsubsec:app_instruct_results}
Table~\ref{tab:gedit3d_instruct} separates direct instruction editing from the agentic pipeline in Sec.~\ref{subsubsec:exp_details}.
Alchemy3D-Instruct outperforms Steer3D on every editing type that Steer3D supports. The margin is largest on addition, where Steer3D has zero success, and on identity preservation, where Steer3D remains near 30--50 while Alchemy3D-Instruct remains above 65. The agentic variant edits one selected view and then applies image-conditioned Alchemy3D, which further raises success: 75.0\% versus 57.5\% on addition, 87.1\% versus 78.1\% on replacement, 52.5\% versus 30.0\% on local appearance, 50.0\% versus 6.7\% on global appearance, and 70.0\% versus 46.7\% on animation. Global appearance remains the weakest setting for the instruction model, consistent with noise in the recaptioned instructions and with the task conflict in Sec.~\ref{subsec:app_future}. Identity preservation is sometimes higher for Alchemy3D-Instruct than for the agentic pipeline, because the intermediate 2D edit changes the asset more strongly.

\paragraph{Multi-turn Editing.}
\begin{figure}[t]
\centering
\includegraphics[width=0.9\linewidth]{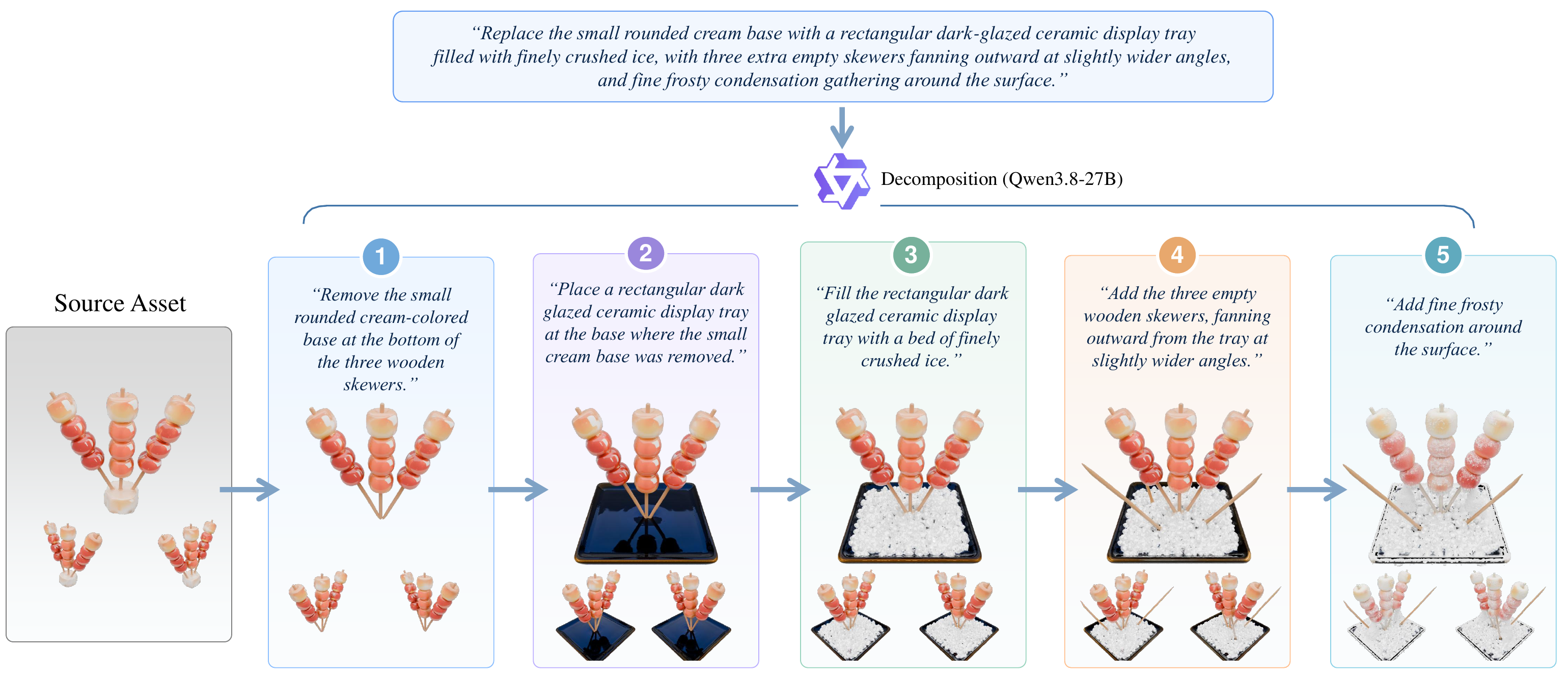}
\caption{Examples of multi-turn, long-horizon 3D asset editing.}
\label{fig:multiturn}
\end{figure}

Fig.~\ref{fig:multiturn} illustrates multi-turn, long-horizon editing with Alchemy3D. A language model (Qwen3.8-27B) decomposes a complex instruction into short atomic edits. The agentic pipeline in Sec.~\ref{subsec:app_eval}, which selects a view, edits that image, and then edits the 3D asset, is applied recursively to each atomic instruction. 

\subsection{Limitations and Future Work}
\label{subsec:app_future}



\paragraph{Imperfect verification.} Although Alchemy3D-1M (Sec.~\ref{sec:dataset}) is a million-scale 3D asset editing dataset, and training Alchemy3D on it supports its utility, several limitations remain. As with prior datasets in image, video, and 3D asset editing~\citep{anyedit, goku,nano3d,3deditformer,partflow} that verify and caption data automatically with vision-language models, our dataset may contain low-quality samples because current VLMs are imperfect. For example, they can struggle to distinguish the left and right sides of an asset, and may accept low-quality samples as valid training pairs.

\paragraph{Task conflict.} In this work, we follow TRELLIS.2~\citep{trellis2} in adopting a dense cross-attention transformer architecture. However, during training, we found that the goals and granularities of different editing types are not always mutually beneficial when trained together. For example, local and global appearance changes are learned poorly when trained jointly with a large number of geometry-changing edits. A mixture-of-experts model~\citep{moe} may separate editing types that operate at different granularities.

\begin{figure}[h]
\begin{center}
\includegraphics[width=\linewidth]{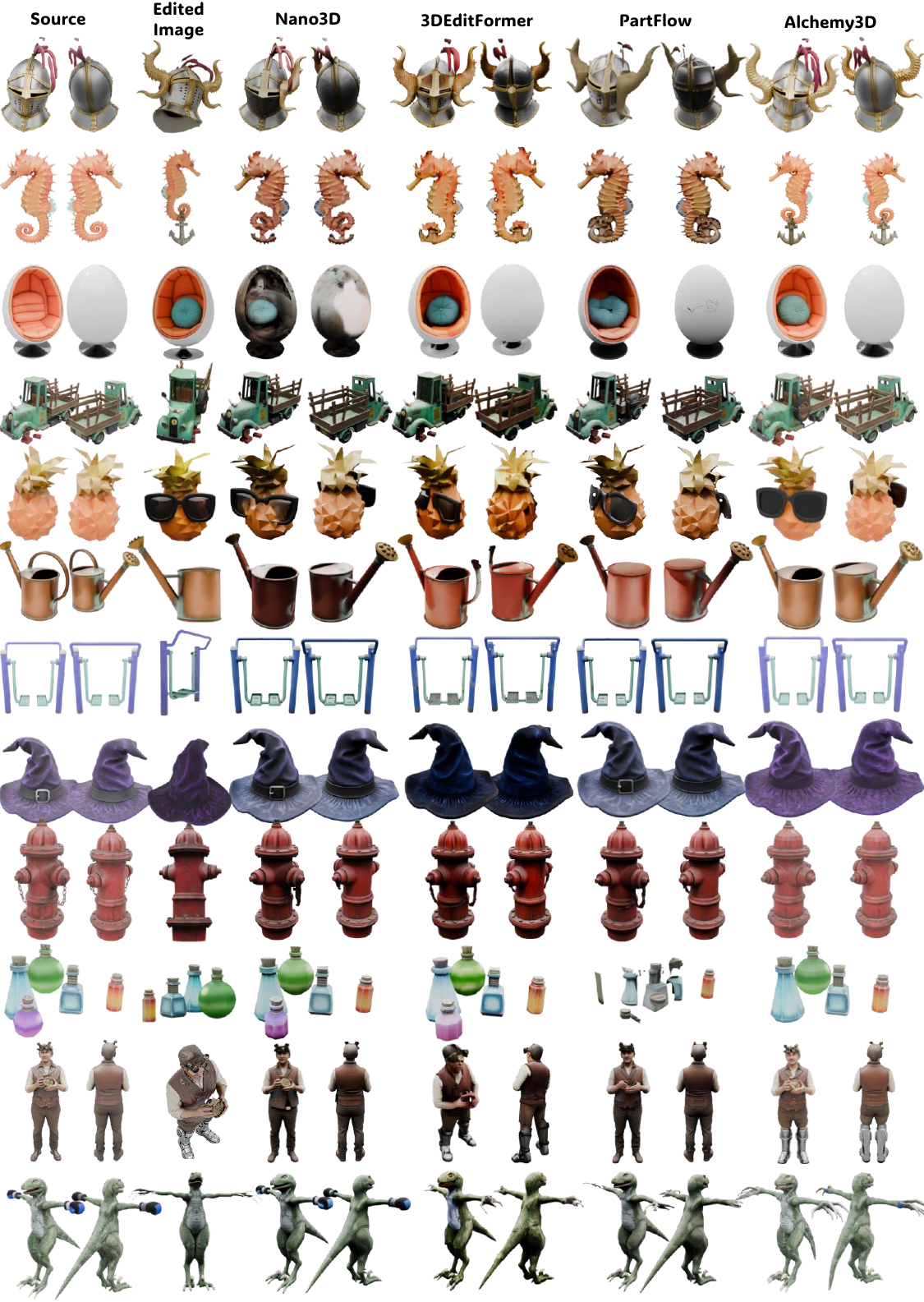}
\end{center}
\caption{Qualitative comparison on addition, removal, and replacement edits.}
\label{fig:qualitative_app}
\end{figure}

\begin{figure}[h]
\begin{center}
\includegraphics[width=\linewidth]{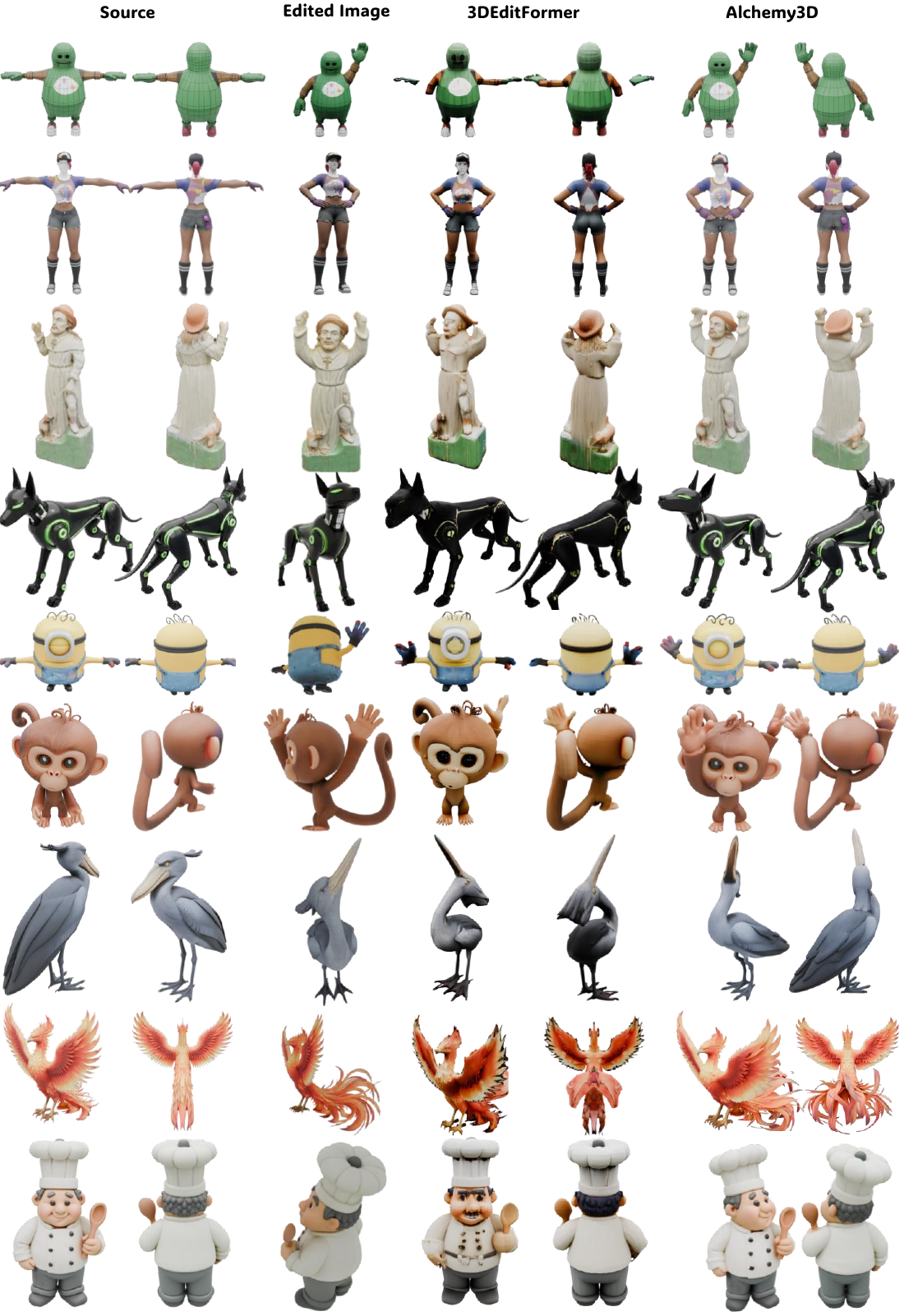}
\end{center}
\caption{Qualitative comparison on animation edits.}
\label{fig:qualitative_animation}
\end{figure}

\begin{figure}[h]
\begin{center}
\includegraphics[width=\linewidth]{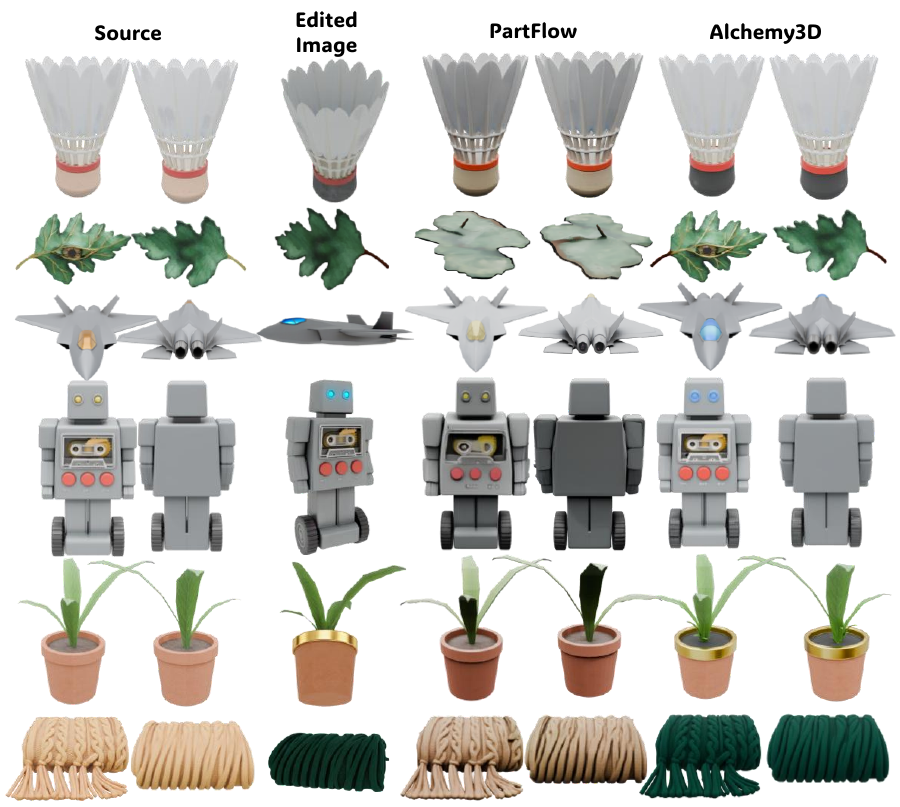}
\end{center}
\caption{Qualitative comparison on appearance edits.}
\label{fig:qualitative_tex}
\end{figure}

\end{document}